\documentclass[12pt,a4paper]{article}

\usepackage{times}
\usepackage{url}
\usepackage{amsmath}
\usepackage{graphicx}
\usepackage{color}
\usepackage{caption}
\usepackage{subcaption}
\usepackage{multirow}
\usepackage{url}
\usepackage{authblk}

\usepackage{algorithmic}
\usepackage{algorithm}

\definecolor{darkred}{rgb}{1.0,0,0}

\makeatletter
\def\@maketitle{%
  \newpage
  \null
  \vskip 2em%
  \begin{center}%
  \let \footnote \thanks
    {\Large\bfseries \@title \par}%
    \vskip 1.5em%
    {\normalsize
      \lineskip .5em%
      \begin{tabular}[t]{c}%
        \@author
      \end{tabular}\par}%
    \vskip 1em%
    {\normalsize \@date}%
  \end{center}%
  \par
  \vskip 1.5em}
\makeatother

\title{Interpreting random forest classification models using a feature contribution method}

{\tiny{
\author[1]{Anna Palczewska\thanks{a.m.wojak@bradford.ac.uk}} 
\author[2]{Jan Palczewski\thanks{j.palczewski@leeds.ac.uk} }
\author[3]{ Richard Marchese Robinson\thanks{r.l.marcheserobinson@ljmu.ac.uk}}
\author[1]{ Daniel Neagu\thanks{d.neagu@bradford.ac.uk}}
\affil[1]{Department of Computing, University of Bradford, BD7 1DP Bradford, UK}
\affil[2]{School of Mathematics, University of Leeds, LS2 9JT Leeds, UK}
\affil[3]{School of Pharmacy and Biomolecular Sciences, \\Liverpool John Moores University, L3 3AF Liverpool, UK}
\date{}
}}

\begin{document}

\maketitle

\begin{abstract}

Model interpretation is one of the key aspects of the model evaluation process. The explanation of the relationship between model variables and outputs is relatively easy for statistical models, such as linear regressions, thanks to the availability of model parameters and their statistical significance. For ``black box'' models, such as random forest, this information is hidden inside the model structure. This work presents an approach for computing feature contributions for random forest classification models. It allows for the determination of the influence of each variable on the model prediction for an individual instance.
By analysing feature contributions for a training dataset, the most significant variables can be determined and their typical contribution towards predictions made for individual classes, i.e., class-specific feature contribution "patterns",  are discovered. These patterns represent a standard behaviour of the model and allow for an additional assessment of the model reliability for a new data. Interpretation of feature contributions  for two UCI benchmark datasets shows the potential of the proposed methodology. The robustness of results is demonstrated through an extensive analysis of feature contributions calculated for a large number of generated random forest models.
\end{abstract}

\section{Introduction}

Models are used to discover interesting patterns in data or  to predict a specific outcome, such as drug toxicity, client shopping purchases, or car insurance premium. They are often used  to support human decisions in various business strategies. This is why it is important to ensure model quality and to understand its outcomes.  Good practice of model development~\cite{Tropsha:2010}  involves: 1) data analysis 2) feature selection, 3) model building and 4) model evaluation.
Implementing these steps together with capturing information on how the data was harvested, how the model was built and how the model was validated, allows us to trust that the model gives reliable predictions. But, how to interpret an existing model? How to analyse the relation between predicted values and the training dataset? Or which features contribute the most to classify a specific instance?

Answers to these questions are considered particularly valuable in such domains as chemoinformatics, bioinformatics or predictive toxicology \cite{Rosenbaum2011}. Linear models, which assign instance-independent coefficients to all features, are the most easily interpreted.
However, in the recent literature, there has been considerable focus on interpreting predictions made by non-linear models do not render themselves to straightforward methods for the determination of variable/feature influence.
In~\cite{Carlsson2009}, the authors present a method for a local interpretation of Support Vector Machine (SVM) and Random Forest models by retrieving the variable corresponding to the largest component of the decision-function gradient at any point in the model. Interpretation of classification models using local gradients is discussed in \cite{Baehrens2010}.  A method for visual interpretation of kernel-based prediction models is described in \cite{Hansen2011}. Another approach, which is presented in detail later, was proposed in \cite{Kuzmin:2011} and aims at shedding light at decision-making process of regression random forests.

Of interest to this paper is a popular ``black-box" model -- the random forest model \cite{Breiman:2001}. Its author suggests two measures of the significance of a particular variable \cite{Breiman:2008}: the variable importance and the Gini importance. The variable importance is derived from the loss of accuracy of model predictions when values of one variable are permuted between instances. Gini importance is calculated from the Gini impurity criterion used in the growing of trees in the random forest. However, in~\cite{Strobl2008}, the authors
showed that the above measures are biased in favor of continuous variables and variables with many categories. They  also demonstrated that the general representation of variable importance is often insufficient for the complete understanding of the relationship between input variables and the predicted value.

Following the above observation, Kuzmin et~al. propose in \cite{Kuzmin:2011} a new technique to calculate a feature contribution, i.e., a contribution of a variable to the prediction, in a random forest model. Their method applies to models generated for data with numerical observed values (the observed value is a real number).  Unlike in the variable importance measures \cite{Breiman:2008}, feature contributions are computed separately for each instance/record. They provide detailed information about relationships between variables and the predicted value. It is the extent and the kind of influence (positive/negative) of a given variable. This new approach was positively tested in~\cite{Kuzmin:2011} on a Quantitative Structure-Activity (QSAR) model for chemical compounds. The results were not only informative about the structure of the model but also provided valuable information for the design of new compounds.

The procedure from~\cite{Kuzmin:2011} for the computation of feature contributions applies to random forest models predicting numerical observed values. This paper aims to extend it to random forest models with categorical predictions, i.e., where the observed value determines one from a finite set of classes. The difficulty of achieving this aim lies in the fact that a discrete set of classes does not have the algebraic structure of real numbers which the approach presented in~\cite{Kuzmin:2011} relies on. Due to the high-dimensionality of the calculated feature contributions, their direct analysis is not easy. 
We develop three techniques for discovering class-specific feature contribution "patterns" in the decision-making process of random forest models: the analysis of median feature contributions, of clusters and log-likelihoods. This facilitates interpretation of model predictions as well as allows a more detailed analysis of model reliability for an unseen data.

The paper is organised as follows. Section~\ref{sec:rf} provides a brief description of random forest models. Section~\ref{sec:fc_binary} presents our approach  for calculating feature contributions for binary classifiers, whilst Section \ref{sec:fc_general} describes its extension to multi-class classification problems. Section~\ref{sec:interpret} introduces three techniques for finding patterns in feature contributions and linking them to model predictions. Section~\ref{sec:application} contains applications of the proposed methodology to two real world datasets from the UCI Machine Learning repository. Section~\ref{sec:conclusions} concludes the work presented in this paper.

\section{Random Forest}\label{sec:rf}

A random forest (RF) model introduced by Breiman \cite{Breiman:2001} is a collection of tree predictors. Each tree is grown according to the following procedure~\cite{Breiman:2008}:
\begin{enumerate}
\item the bootstrap phase: select randomly a subset of the training dataset -- a local training set for growing the tree. The remaining samples in the training dataset form a so-called out-of-bag (OOB) set and are used to estimate the RF's goodness-of-fit.
\item the growing phase: grow the tree by splitting the local training set at each node according to the value of one variable from a randomly selected subset of variables (the best split) using classification and regression tree (CART) method \cite{Breiman:1984}.
\item each tree is grown to the largest extent possible. There is no pruning.
\end{enumerate}
The bootstrap and  growing phases require an input of random quantities. It is assumed that these quantities are independent between trees and identically distributed. Consequently, each tree can be viewed as sampled independently from the ensemble of all tree predictors for a given training dataset.

For prediction, an instance is run through each tree in a forest down to a terminal node which assigns it a class. Predictions supplied by the trees undergo a voting process: the forest returns ca class with the maximum number of votes. Draws are resolved through a random selection.

To present our feature contribution procedure in the following section, we have to develop a probabilistic interpretation of the forest prediction process. Denote by $C = \{ C_1, C_2, \ldots, C_K \}$ the set of classes and by $\Delta_K$ the set
\[
\Delta_K = \big\{ (p_1, \ldots, p_K): \ \sum_{k=1}^K p_k = 1 \text{ and } p_k \ge 0 \big\}.
\]
An element of $\Delta_K$ can be interpreted as a probability distribution over $C$. Let $e_k$ be an element of $\Delta_K$ with $1$ at position $k$ -- a probability distribution concentrated at class $C_k$. If a tree $t$ predicts that an instance $i$ belongs to a class $C_k$ then we write $\hat{Y}_{i,t} = e_k$. This provides a mapping from predictions of a tree to the set $\Delta_K$ of probability measures on $C$. Let
\begin{equation}\label{eqn:hatY}
\hat{Y}_i=\frac{1}{T}\sum_{t=1}^{T}\hat{Y}_{i,t},
\end{equation}
where $T$  is the overall number of trees in the forest. Then $\hat Y_i \in \Delta_K$ and the prediction of the random forest for the instance $i$ coincides with a class $C_k$ for which the $k$-th coordinate of $\hat Y_i$ is maximal.\footnote{The distribution $\hat Y_i$ is calculated by the function \texttt{predict} in the R package \texttt{randomForest} \cite{Liaw:2002} when the type of prediction is set to \texttt{prob}.}

\section{Feature Contributions for Binary Classifiers}\label{sec:fc_binary}

The set $\Delta_K$ simplifies considerably when there are two classes, $K=2$. An element $p \in \Delta_K$ is uniquely represented by its first coordinate $p_1$ ($p_2 = 1-p_1$). Consequently, the set of probability distributions on $C$ is equivalent to the probability weight assigned to class $C_1$.

Before we  present our method for computing feature contributions, we have to examine the tree growing process. After selecting a training set, it is positioned in the root node. A splitting variable (feature) and a splitting value are selected and the set of instances is split between the left and the right child of the root node. The procedure is repeated until all instances in a node are in the same class or further splitting does not improve prediction. The class that a tree assigns to a terminal node is determined through majority voting between instances in that node.

We will refer to instances of the local training set that pass through a given node as the training instances in this node. The fraction of the training instances in a node $n$ belonging to class $C_1$ will be denoted by $Y^n_{mean}$. This is the probability that a randomly selected element from the training instances in this node is in the first class. In particular, a terminal node is assigned to class $C_1$ if $Y^n_{mean} > 0.5$ or $Y^n_{mean} = 0.5$ and the draw is resolved in favor of class $C_1$.

The feature contribution procedure for a given instance involves two steps: 1) the calculation of local increments of feature contributions for each tree and 2) the aggregation of feature contributions over the forest. A local increment corresponding to a feature $f$ between a parent node (p) and a child node (c) in a tree is defined as follows:
\[
LI_f^c =
\begin{cases}
Y_{mean}^{c}-Y_{mean}^{p},& \parbox{4.05cm}{if the split in the parent is performed over the feature $f$,}\\[10pt]
0, & \text{otherwise.}
\end{cases}
\]
A local increment for a feature $f$ represents the change of the probability of being in class $C_1$ between the child  node and its parent node provided that $f$ is the splitting feature in the parent node.  It is easy to show that the sum of these changes, over all features, along the path followed by an instance from the root node to the terminal node in a tree is equal to the difference between $Y_{mean}$ in the terminal and the root node.

The contribution $FC^f_{i, t}$ of a feature $f$ in a tree $t$ for an instance $i$ is equal to the sum of $LI_f$ over all nodes on the path of instance $i$ from the root node to a terminal node. The contribution of a feature $f$ for an instance $i$ in the forest is then given by
\begin{equation}\label{eqn:feat:contrib}
FC^f_{i}=\frac{1}{T}\sum_{t=1}^TFC^f_{i,t}.
\end{equation}
The feature contributions vector for an instance $i$ consists of contributions $FC^f_i$ of all features $f$.

Notice that if the following condition is satisfied:
\begin{itemize}
\item[\textbf{(U)}] for every tree in the forest, local training instances in each terminal node are of the same class
\end{itemize}
then $\hat Y_i$ representing forest's prediction \eqref{eqn:hatY} can be written as
\begin{equation}\label{eqn:total_sum}
\hat Y_i = \Big( Y^{r} + \sum_{f} FC^f_i,\ 1- Y^{r} - \sum_{f} FC^f_i \Big)
\end{equation}
where $Y^{r}$ is the coordinate-wise average of $Y_{mean}$ over all root nodes in the forest. If this unanimity condition (U) holds, feature contributions can be used to retrieve predictions of the forest. Otherwise, they only allow for the interpretation of the model.

\subsection{Example}

\begin{figure}[tbh]\centering
6\includegraphics[width=\textwidth]{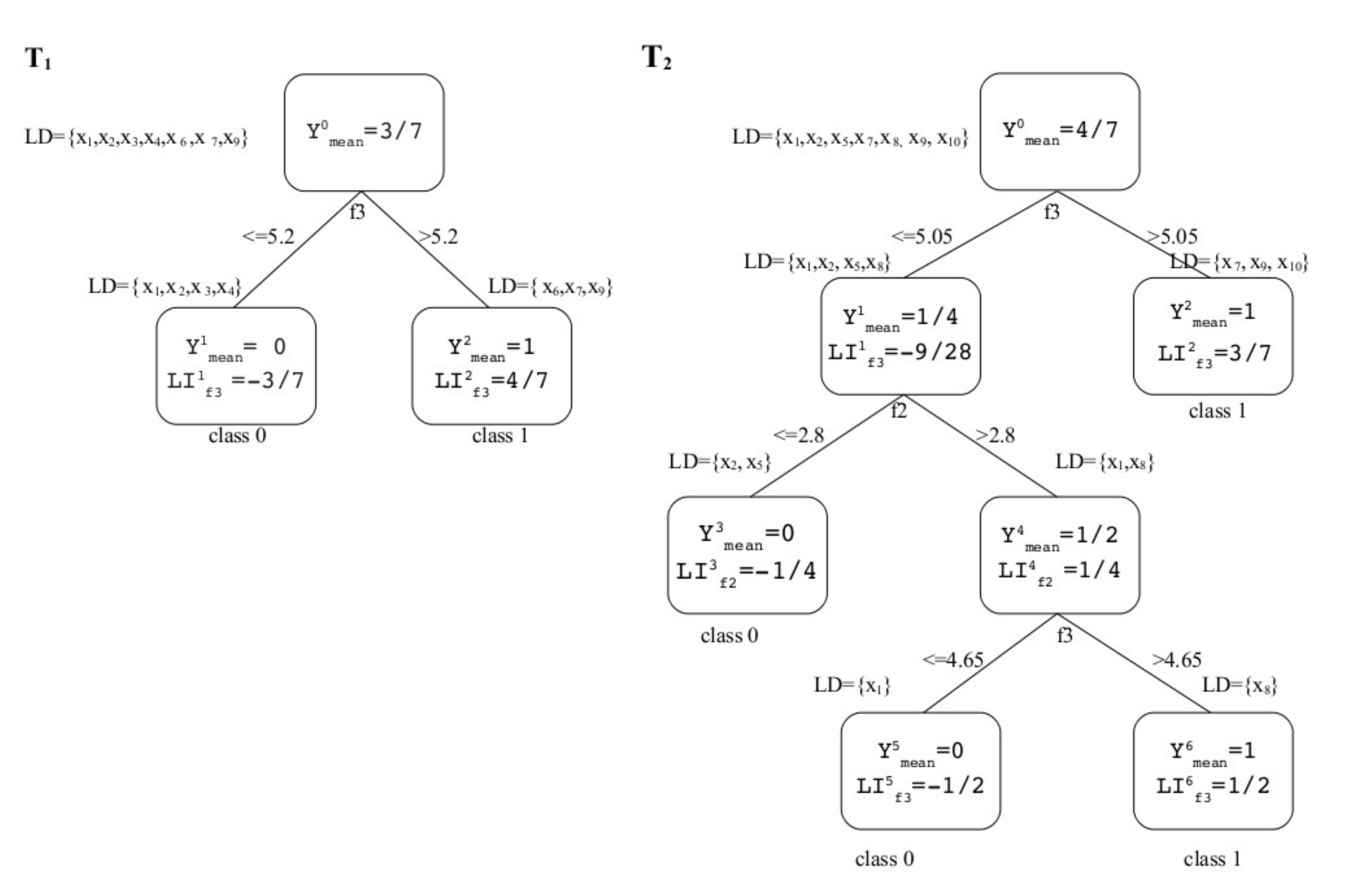}
   \caption{A random forest model for the dataset from Table \ref{table:example_bfc_data}. The set LD in the root node contains a local training set for the tree. The sets LD in the child nodes correspond to the split of the above set according to the value of selected feature. In each node, $Y^n_{mean}$ denotes the fraction of instances in the LD set in this node belonging to class 1, whilst $LI^n_f$ shows non-zero local increments.}
\label{Fig:featureContribution}
\end{figure}

We will demonstrate the calculation of feature contributions on a toy example using a subset of the UCI Iris Dataset~ \cite{Iris}. From the original dataset, ten records were selected -- five for each of two types of the iris plant: versicolor (class 0) and virginica (class 1) (see Table~\ref{table:example_bfc_data}).  A plant is represented by four attributes: Sepal.Length (f1),  Sepal.Width (f2), Petal.Length (f3) and  Petal.Width (f4). This dataset was used to generate a random forest model with two trees, see Figure~\ref{Fig:featureContribution}.  In each tree, the local training set (LD) in the root node collects those records which were chosen by the random forest algorithm to build that tree. The LD sets in the child nodes correspond to the split of the above set according to the value of a selected feature (it is written between branches). This process is repeated until reaching terminal nodes of the tree. Notice that the condition \textbf{(U)} is satisfied --  for both trees, each terminal node contains local training instances of the same class: $Y_{mean}$ is either $0$ or $1$.

\begin{table}[tbh]\centering
\caption{Selected records from the UCI Iris Dataset. Each record corresponds to a plant.  The plants were classified as iris versicolor (class 0) and virginica (class 1).}
\label{table:example_bfc_data}
\begin{tabular}{c|cc|cc|c}
\hline
&\multicolumn{2}{c|}{Sepal}&\multicolumn{2}{|c|}{Petal}&\\
 &  Length (f1) &  Width (f2)  &  Length (f3) &  Width (f4) & class\\
\hline
$x_1$  & 6.4 & 3.2 & 4.5 & 1.5 & {\bf0} \\
$x_2$  & 6.3 & 2.5 & 4.9 & 1.5 & {\bf 0} \\
$x_3$  & 6.4 & 2.9 & 4.3 & 1.3 & {\bf 0} \\
$x_4$  & 5.5 & 2.5 & 4.0 & 1.3 & {\bf 0} \\
$x_5$  & 5.5 & 2.6 & 4.4 & 1.2 & {\bf 0} \\
$x_6$  & 7.7 & 3.0 & 6.1 & 2.3 & {\bf 1} \\
$x_7$  & 6.4 & 3.1 & 5.5 & 1.8 & {\bf 1} \\
$x_8$  & 6.0 & 3.0 & 4.8 & 1.8 & {\bf 1} \\
$x_9$  & 6.7 & 3.3 & 5.7 & 2.5 & {\bf 1} \\
$x_{10}$ & 6.5 & 3.0 & 5.2 & 2.0 & {\bf 1} \\
\hline
\end{tabular}
\end{table}

The process of calculating feature contributions runs in 2 steps: the determination of local increments for each node in the forest (a preprocessing step) and the calculation of feature contributions for a particular instance. Figure~\ref{Fig:featureContribution} shows $Y^n_{mean}$ and the local increment $LI_f^c$ for a splitting feature $f$ in each node. Having computed these values, we can calculate feature contributions for an instance by running it through both trees and summing local increments of each of the four features. For example, the contribution of a given feature for the instance $x_1$ is calculated by summing local increments for that feature along the path $p_1=n_0 \rightarrow n_1$ in tree $T_1$  and the path $p_2=n_0\rightarrow n_1 \rightarrow n_4\rightarrow n_5$ in tree {\bf $T_2$}.  According to Formula~\eqref{eqn:feat:contrib} the contribution of feature f2 is calculated as
$$
FC_{x_1}^{f2}= \frac{1}{2}\Big(0+\frac{1}{4}\Big) = 0.125
$$
and the contribution of feature f3 is
$$
FC_{x_1}^{f3}= \frac{1}{2}\Big(-\frac{3}{7}-\frac{9}{28}-\frac{1}{2}\Big)=-0.625.
$$
The contributions of features f1 and f4 are equal to $0$ because these attributes are not used in any decision made by the forest. The predicted probability $\hat Y_{x_1}$ that $x_1$ belongs to class 1 (see Formula~\eqref{eqn:total_sum}) is
\[
\hat{Y}_{x_1}= \underbrace{\frac{1}{2}\Big(\frac{3}{7}+\frac{4}{7}\Big)}_{\hat Y^r} + \underbrace{\big(0 + 0.125 -0.625+0\big)}_{\sum_{f} FC^f_{x_1}}=0.0
\]

Table~\ref{table:fc} collects feature contributions for all 10 records in the example dataset. These results can be interpreted as follows:
\begin{itemize}
 \item for instances $x_1, x_3$, the contribution of f2 is positive, i.e., the value of this feature increases the probability of being in class 1 by 0.125. However, the large negative contribution of the feature f3 implies that the value of this feature for instances $x_1$ and $x_3$ was decisive in assigning the class 0 by the forest.
\item for instances $x_6, x_7, x_9$,  the decision is based only on the feature f3.
\item for instances $x_2, x_4, x_5$, the contribution of both features leads the forest decision towards class 0.
\item for instances $x_8$, $x_{10}$, $\hat Y$ is $0.5$. This corresponds to the case where one of the trees points to class 0 and the other to class 1. In practical applications, such situations are resolved through a random selection of the class. Since $\hat Y^r = 0.5$, the lack of decision of the forest has a clear interpretation in terms of feature contributions: the amount of evidence in favour of one class is counterbalanced by the evidence pointing towards the other.
\end{itemize}

\begin{table}[t!!]\centering
\caption{Feature contributions for the random forest model from Figure \ref{Fig:featureContribution}.}
\label{table:fc}
\begin{tabular}{c|c|cc|cc|c}
\hline

&&\multicolumn{2}{c|}{Sepal}&\multicolumn{2}{|c|}{Petal}&\\
& $\hat{Y}$ & Length (f1) & Width (f2) & Length (f3) & Width (f4) & prediction\\
\hline
$x_1$ & 0.0 & 0 &  0.125 & -0.625 & 0 & {\bf0} \\
$x_2$ & 0.0 & 0 & -0.125 & -0.375 & 0 & {\bf 0} \\
$x_3$ & 0.0 & 0 &  0.125 & -0.625 & 0 & {\bf 0} \\
$x_4$ & 0.0 & 0 & -0.125 & -0.375 & 0 & {\bf 0} \\
$x_5$ & 0.0 & 0 & -0.125 & -0.375 & 0 & {\bf 0}\\
$x_6$ & 1.0 & 0 &  0 	 & 0.5 	  & 0 & {\bf 1} \\
$x_7$ & 1.0 & 0 &  0     & 0.5    & 0 & {\bf 1} \\
$x_8$ & 0.5 & 0 &  0.125 & -0.125  & 0 & {\bf ?} \\
$x_9$ & 1.0 & 0 &  0     & 0.5    & 0 & {\bf 1} \\
$x_{10}$& 0.5 & 0 &  0     & 0      & 0 & {\bf ?} \\
\hline
\end{tabular}

\end{table}

\section{Feature Contributions for General Classifiers} \label{sec:fc_general}

When $K > 2$, the set $\Delta_K$ cannot be described by a one-dimensional value as above. We, therefore, generalize the quantities introduced in the previous section to a multi-dimensional case. $Y^n_{mean}$ in a node $n$ is an element of $\Delta_K$, whose $k$-th coordinate, $k=1, 2, \ldots, K$, is defined as
\begin{equation}\label{eqn:Y_mean}
Y^n_{mean, k} = \frac{|\{i \in TS(n):\ i \in C_k\}|}{|TS(n)|},
\end{equation}
where $TS(n)$ is the set of training instances in the node $n$ and $|\cdot|$ denotes the number of elements of a set.
Hence, if an instance is selected randomly from a local training set in a node $n$, the probability that this instance is in class $C_k$ is given by the $k$-th coordinate of the vector $Y^n_{mean}$. Local increment $LI^c_f$ is analogously generalized to a multidimensional case:
\[
LI_f^c =
\begin{cases}
Y_{mean}^{c}-Y_{mean}^{p},& \parbox{4.05cm}{if the split in the parent is performed over the feature $f$,}\\[10pt]
\underbrace{(0, \ldots, 0)}_{\text{$K$ times}}, & \text{otherwise,}
\end{cases}
\]
where the difference is computed coordinate-wise. Similarly, $FC_{i,t}^f$ and $FC_i^f$ are extended to vector-valued quantities.
Notice that if the condition (U) is satisfied, Equation \eqref{eqn:total_sum} holds with $Y^r$ being a coordinate-wise average of vectors $Y^n_{mean}$ over all root nodes $n$ in the forest.

Take an instance $i$ and let $C_k$ be the class to which the forest assigns this instance. Our aim is to understand which variables/features drove the forest to make that prediction. We argue that the crucial information is the one which explains the value of the $k$-th coordinate of $\hat Y_i$. Hence, we want to study the $k$-th coordinate of $FC^f_i$ for all features $f$.

Pseudo-code to calculate feature contributions for a particular instance towards the class predicted by the random forest is presented in Algorithm~\ref{alg:feature_contribution}. Its inputs consist of a random forest model $RF$ and an instance $i$ which is represented as a vector of feature values. In line~1, $k \in \{1, 2, \ldots, K\}$ is assigned the index of a class predicted by the random forest RF  for the instance $i$. The following line creates a vector of real numbers indexed by features and initialized to $0$. Then for each tree in the forest $RF$ the instance $i$ is run down the tree and feature contributions are calculated. The quantity $SplitFeature(parent)$ identifies a feature $f$ on which the split is performed in the node $parent$. If the value $i(f)$ of that feature $f$ for the instance $i$ is lower or equal to the threshold $SplitValue(parent)$, the route continues to the left child of the node $parent$. Otherwise, it goes to the right child (each node in the tree has either two children or is a terminal node). A position corresponding to the feature $f$ in the vector $FC$ is updated according to the change of value of $Y_{mean,k}$, i.e., the $k$-th coordinate of $Y_{mean}$, between the parent and the child.

\begin{algorithm}[tbh!!]
\caption{FC($RF$,$i$)}
\label{alg:feature_contribution}
\begin{algorithmic}[1]
\STATE $k \leftarrow forest\_predict(RF, i)$
\STATE $FC \leftarrow vector(features)$
  \FOR{each tree $T$ in forest $F$}
   \STATE $parent \leftarrow root(T)$
   \WHILE{$parent\ !=\ \text{TERMINAL}$}
	  \STATE{$f \leftarrow SplitFeature(parent)$}
          \IF{$i[f] <= SplitValue (parent)$}
	    \STATE $child \leftarrow leftChild(parent)$
	    \ELSE
             \STATE $child \leftarrow rightChild(parent)$
	    \ENDIF
	    \STATE $FC[f] \leftarrow FC[f] + Y_{mean, k}^{child} - Y_{mean, k}^{parent}$
	\STATE $parent \leftarrow child$
	\ENDWHILE
  \ENDFOR
  \STATE $FC \leftarrow FC $ / nTrees($F$)
\RETURN $FC$
\end{algorithmic}
\end{algorithm}

 Algorithm~\ref{alg:local_increment} provides a sketch of the preprocessing step to compute $Y^n_{mean}$ for all nodes $n$ in the forest. The parameter $D$ denotes the set of instances used for training of the forest $RF$. In line~2, $TS$ is assigned the set used for growing tree $T$. This set is further split in nodes according to values of splitting variables. We propose to use DFS (depth first search \cite{Cormen:2001}) to traverse the tree and compute the vector $Y^n_{mean}$ once a training set for a node $n$ is determined. There is no need to store a training set for a node $n$ once $Y^n_{mean}$ has been calculated.

\begin{algorithm}[tbh!!]
\caption{$Y_{mean}(RF, D)$}
\label{alg:local_increment}
\begin{algorithmic}[1]
\FOR{each tree $T$ in forest $F$}
\STATE $TS \leftarrow$ training set for tree $T$
\STATE use DFS algorithm to compute training sets in all other nodes $n$ of tree $T$ and compute the vector $Y^n_{mean}$ according to formula \eqref{eqn:Y_mean}.
\ENDFOR
\end{algorithmic}
\end{algorithm}
\section{Analysis of Feature Contributions} \label{sec:interpret}

Feature contributions provide the means to understand mechanisms that lead the model towards particular predictions. This is important in chemical or biological applications where the additional knowledge of the forest's decision-making process can inform the development of new chemical compounds or explain their interactions with living organisms. Feature contributions may also be useful for assessing the reliability of model predictions for unseen instances. They provide complementary information to forest's voting results.
This section proposes three techniques for finding patterns in the way a random forest uses available features and linking these patterns with the forest's predictions.

\subsection{Median}
The median of a sequence of numbers is such a value that the number of elements bigger than it and the number of elements smaller than it is identical. When the number of elements in the sequence is odd, this is the central elements of the sequence. Otherwise, it is common to take the midpoint between the two most central elements. In statistics, the median is an estimator of the expectation which is less affected by outliers than the sample mean. We will use this property of the median to find a ``standard level'' of feature contributions for representatives of a particular class. This standard level will facilitate an understanding of which features are instrumental for the classification. It can also be used to judge the reliability of forest's prediction for an unseen instance.

For a given random forest model, we select those instances from the training dataset that are classified correctly. We calculate the medians of contributions of every feature separately for each class. The medians computed for one class are combined into a vector which is interpreted as providing the aforementioned ``standard level'' for this class. If most of instances from the training dataset belonging to a particular class are close to the corresponding vector of medians, we may treat this vector justifiably as a standard level. When a prediction is requested for a new instance, we query the random forest model for the fraction of trees voting for each class and calculate feature contributions leading to its final prediction. If a high fraction of trees votes for a given class and the feature contributions are close to the standard level for this class, we may reasonably rely on the prediction. Otherwise we may doubt the random forest model prediction.

It may, however, happen that many instances from the training dataset correctly predicted to belong to a particular class are distant from the corresponding vector of medians. This might suggest that there is more than one standard level, i.e., there are multiple mechanisms relating features to correct classes.
The next subsection presents more advanced methods capable of finding a number of standard levels -- distinct patterns followed by the random forest model in its prediction process.

\subsection{Cluster Analysis}\label{subsec:clustering}

Clustering is an approach for grouping elements/objects according to their similarity \cite{Hand:2001}.  This allows us to discover patterns that are characteristic for a particular group.  As we discussed above, feature contributions in one class may have more than one "standard level". When this is discovered, clustering techniques can be employed to find if there is a small number of distinct standard levels, i.e., feature contributions of the instances in the training dataset group around a few points with only a relatively few instances being far away from them. These few instances are then treated as unusual representatives of a given class.
We shall refer to clusters of instances around these standard levels as "core clusters".

The analysis of core clusters can be of particular importance for applications. For example, in the classification of chemical compounds, the split into clusters may point to groups of compounds with different mechanisms of activity. We should note that the similarity of feature contributions does not imply that particular features are similar. We examined several examples and noticed that clustering based upon the feature values did not yield useful results whereas the same method applied to feature contributions was able to determine a small number of core clusters.

\begin{figure}[tb]\centering
\includegraphics[scale=0.4]{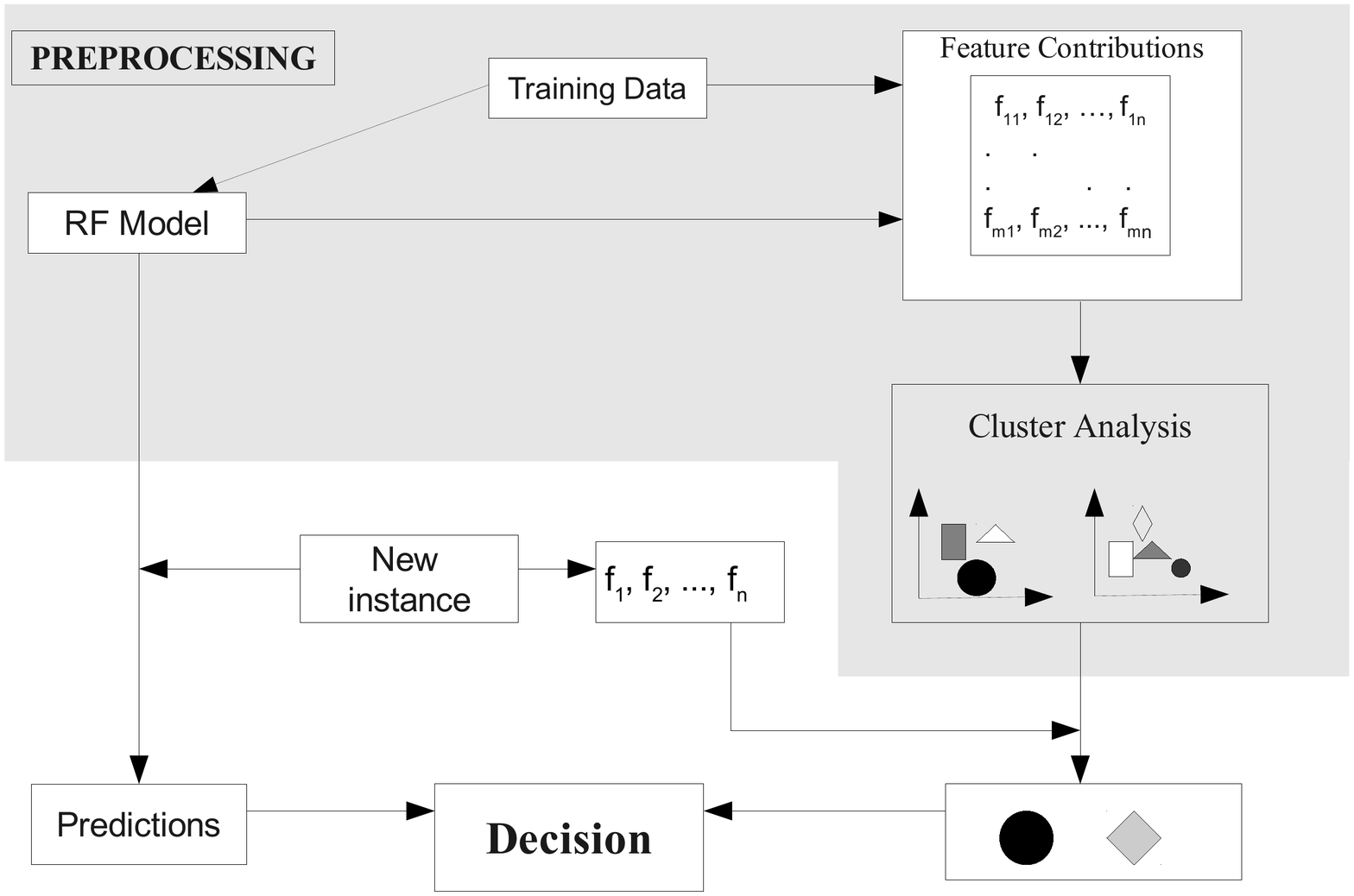}
\caption{The workflow for assessing the reliability of the prediction made by a random forest (RF) model.}
\label{Fig:WorkFlow2}
\end{figure}

Figure \ref{Fig:WorkFlow2}  demonstrates the process of analysis of model reliability for a new instance  using cluster analysis.  In a preprocessing phase, feature contributions for instances in the training dataset are obtained. The optimal number of clusters for each class can be estimated by using, e.g., the Akaike information criterion (AIC), the Bayesian information criterion (BIC) or the Elbow method \cite{Hand:2001,Rajaraman:2012}. We noticed that these methods should not be rigidly adhered to: their underlying assumption is that the data is clustered and we only have to determine the number of these clusters. As we argued above, we expect feature contributions for various instances to be grouped into a small number of clusters and we accept a reasonable number outliers interpreted as unusual instances for a given class. Clustering algorithms try to push those outliers into clusters, hence increasing their number unnecessarily. We recommend, therefore, to treat the calculated optimal number of clusters as the maximum value and consecutively decrease it looking at the structure and performance of the resulting clusters: for each cluster we assess the average fraction of trees voting for the predicted class across the instances belonging to this cluster as well as the average distance from the centre of the cluster. Relatively large clusters with the former value close to $1$ and the latter value small form the group of core clusters.

To assess the reliability of the model prediction for a new instance, we recommend looking at two measures: the fraction of trees voting for the predicted class as well as the cluster to which the instance is assigned based on its feature contributions. If the cluster is one of the core clusters and the distance from its centre is relatively small, the instance is a typical representative of its predicted class. This together with high decisiveness of the forest suggests that the model's prediction should be trusted. Otherwise, we should allow for an increased chance of misclassification.

\subsection{Log-likelihood}

\begin{figure}[tbh]\centering
\includegraphics[width = 0.5\textwidth, angle=270]{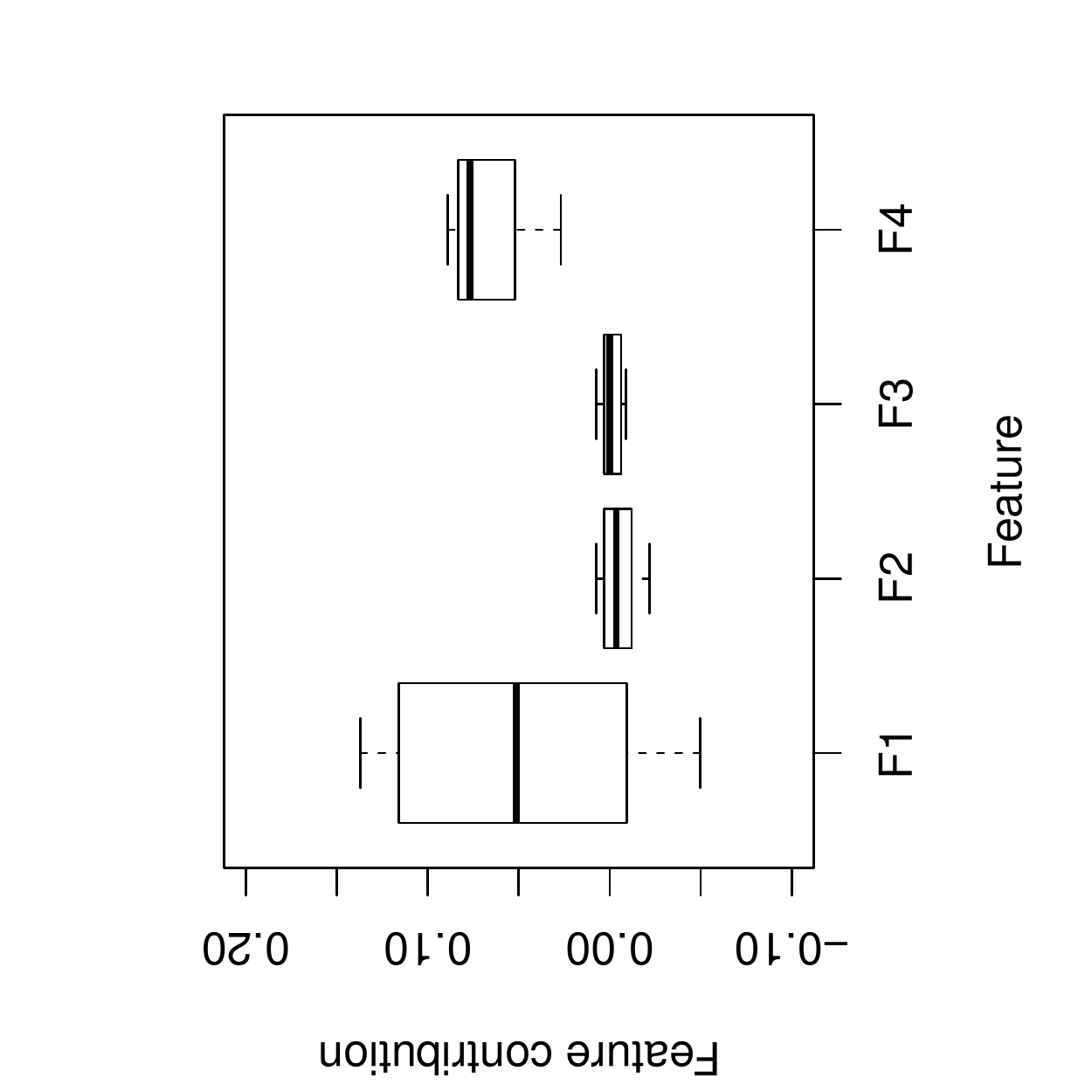}
\caption{The box-plot for feature contributions within a core cluster for a hypothetical random forest model.}
\label{fig:cluster_boxplot_toy}
\end{figure}

Feature contributions for a given instance form a vector in a multi-dimensional Euclidean space.
Using a popular $k$-mean clustering method, for each class we divide vectors corresponding to feature contributions of instances in the training dataset into groups minimizing the Euclidean distance from the centre in each group.
Figure \ref{fig:cluster_boxplot_toy} shows a box-plot of feature contributions for instances in a core cluster in a hypothetical random forest model.
Notice that some features are stable within a cluster -- the height of the box is small. Others (F1 and F4) display higher variability. One would therefore expect that the same divergence of contributions for features F3 and F4 from their mean value should be treated differently. It is more significant for the feature F3 than for the feature F4. This is unfortunately not taken into account when the Euclidean distance is considered. Here, we propose an alternative method for assessing the distance from the cluster centre which takes into account the variation of feature contributions within a cluster. Our method has probabilistic roots and we shall present it first from a statistical point of view and provide other interpretations afterwards.

We assume that feature contributions for instances within a cluster share the same base values $(\mu_f)$ - the centre of the cluster. We attribute all discrepancies between this base value and the actual feature contributions to a random perturbation. These perturbations are assumed to be normally distributed with the mean $0$ and the variance $\sigma_f^2$, where $f$ denotes the feature. The variance of the perturbation for each feature is selected separately -- we use the sample variance computed from feature contributions of instances of the training dataset belonging to this cluster.
Although it is clear that perturbations for different features exhibit some dependence, it is impossible to assess it given the number of instances in a cluster and a large number of features typically in use.\footnote{A covariance matrix of feature contribution has $F(F+1)/2$ distinct entries, where $F$ is the number of features. This value is usually larger than the size of a cluster making it impossible to retrieve useful information about the dependence structure of feature contributions. Application of more advanced methods, such as principal component analysis, is left for future research.} Therefore, we resort to a common solution: we assume that the dependence between perturbations is small enough to justify treating them as independent. Summarising, our statistical model for the distribution of feature contributions within a cluster is as follows: feature contributions for instances within a cluster are composed of a base value and a random perturbation which is normally distributed and independent between features.

Take an instance $i$ with feature contributions $FC_i^f$. The log-likelihood of being in a cluster with the centre $(\mu_f)$ and variances of perturbations $(\sigma^2_f)$ is given by
\begin{equation}\label{eqn:log-like}
LL_i = \sum_{f} \Big( -\frac{(FC_i^f - \mu_f)^2}{2 \sigma^2_f} - \frac12 \log (2\pi \sigma^2_f) \Big).
\end{equation}
The higher the log-likelihood the bigger the chance of feature contributions of the instance $i$ to belong to the  cluster. Notice that the above sum takes into account the observations we made at the beginning of this subsection. Indeed, as the second term in the sum above is independent of the considered instance, the log-likelihood is equivalent to
\[
\sum_{f} \Big( -\frac{(FC_i^f - \mu_f)^2}{2 \sigma^2_f}\Big),
\]
which is the negative of the squared weighted Euclidean distance between $FC_i^f$ and $\mu_f$. The weights being inversely proportional to the variability of a given feature contribution in the training instances in the  cluster. In our toy example of Figure \ref{fig:cluster_boxplot_toy}, this corresponds to penalizing more for discrepancies for features F2 and F3, and significantly less for discrepancies for features F1 and F4.

In the following section, we analyse relations between the log-likelihood and classification for a UCI Breast Cancer Wisconsin Dataset.

\section{Applications}\label{sec:application}

In this section, we demonstrate how the techniques from the previous section can be applied to improve understanding of a random forest model. We consider one example of a binary classifier using the UCI Breast Cancer Wisconsin Dataset \cite{BreastCancer} (BCW Dataset) and one example of a general classifier for the UCI Iris Dataset \cite{Iris}. We complement our studies with a robustness analysis.

\subsection{Breast Cancer Wisconsin Dataset}\label{subsec:empirics_breast}

The UCI Breast Cancer Wisconsin Dataset contains characteristics of cell nuclei for 569 breast tissue samples; 357 are diagnosed as benign and 212 as malignant. The characteristics were captured from a digitized image of a fine needle aspirate (FNA) of a breast mass. There are 30 features, three (the mean, the standard error and the average of the three largest values) for each of the following 10 characteristics: radius, texture,  perimeter, area, smoothness, compactness, concavity, concave points, symmetry and fractal dimension. For brevity, we numbered these features from $F1$ to $F30$ according to their order in the data file.

To reduce correlation between features and facilitate model interpretation, the min-max (minimal-redundancy-maximal-relevance) method was applied and the following features were removed from the dataset: 1, 3, 8, 10, 11, 12, 13, 15, 19, 20, 21, 24, 26. A  random forest model was generated on 2/3 randomly selected instances using 500 trees. The other 1/3 of instances formed the testing dataset. The validation showed that the model accuracy was 0.9682 (only 6 instances out of 189 were classified incorrectly); similar accuracy was achieved when the model was generated using all the features.

\begin{figure}[h!]\centering
\includegraphics[scale=0.4, angle=270]{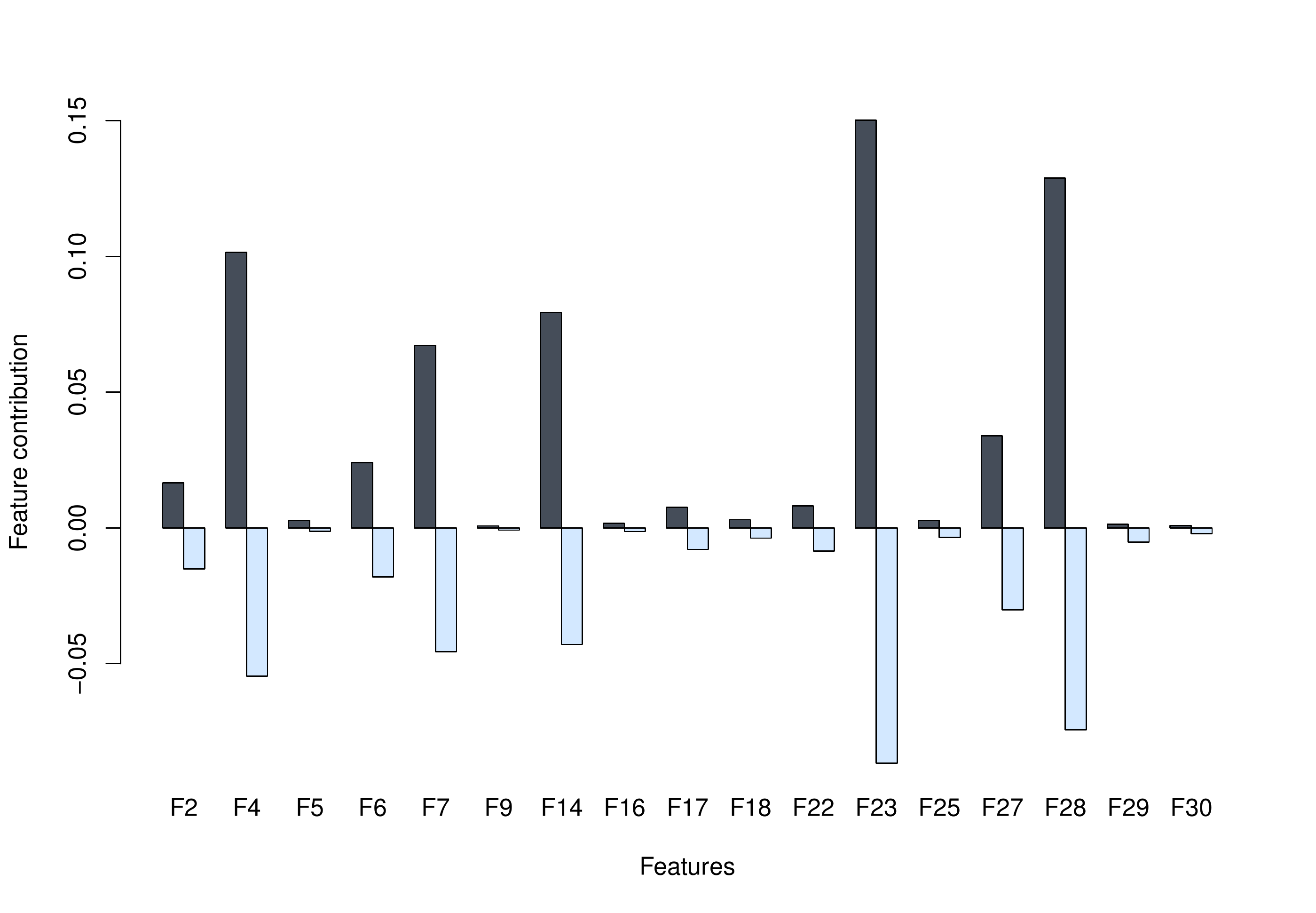}
   \caption{Medians of feature contributions for each class for the BCW Dataset. The light grey bars represent contributions toward class 0 and the black bars show contributions towards class 1.  }
\label{Fig:medians_fc}
\end{figure}

\begin{figure}[h!]\centering
\includegraphics[height=0.8\textwidth, angle=270]{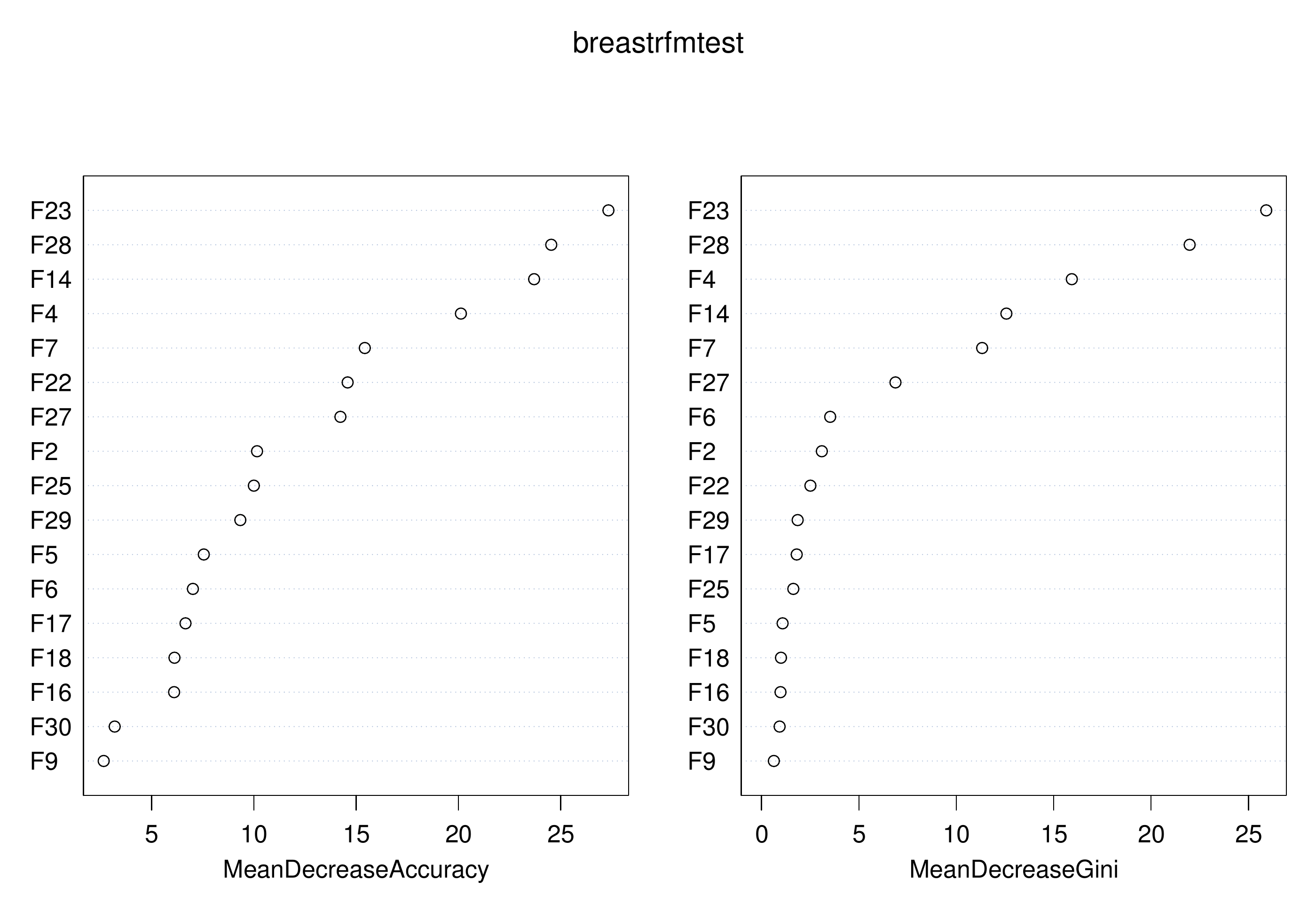}
\caption{The left panel shows permutation based variable importance and the right panel displays Gini importance for a RF binary classification model developed for the BCW Dataset. Graphs generated using randomForest package in R.}
\label{Fig:varImp}
\end{figure}

We applied our feature contribution algorithm to the above random forest binary classifier. To align notation with the rest of the paper, we denote the class ``malignant'' by 1 and the class ``benign" by 0. Aggregate results for the feature contributions for all training instances and both classes are presented in Figure~\ref{Fig:medians_fc}. Light-grey bars show medians of contributions for instances of class 0, whereas black bars show medians of contributions for instances of class 1 (malignant). Notice that there are only a few significant features in the graph: F4 -- the mean of the cell area,  F7 -- the mean of the cell concavity, F14 -- the standard deviation of the cell area, F23 -- the average of three largest measurements of the cell perimeter and F28 -- the average of three largest measurements of concave points. This selection of significant features is perfectly in agreement with the results of the permutation based variable importance (the left panel of Figure \ref{Fig:varImp}) and the Gini importance (the right panel of Figure \ref{Fig:varImp}). Interpreting the size of bars as the level of importance of a feature, our results are in line with those provided by the Gini index. However, the main advantage of the approach presented in this paper lies in the fact that one can study the reasons for the forest's decision for a \emph{particular instance}.

\begin{figure}[p]
\begin{subfigure}[b]{1\textwidth}
\centering
\includegraphics[width=0.6\textwidth, angle=270]{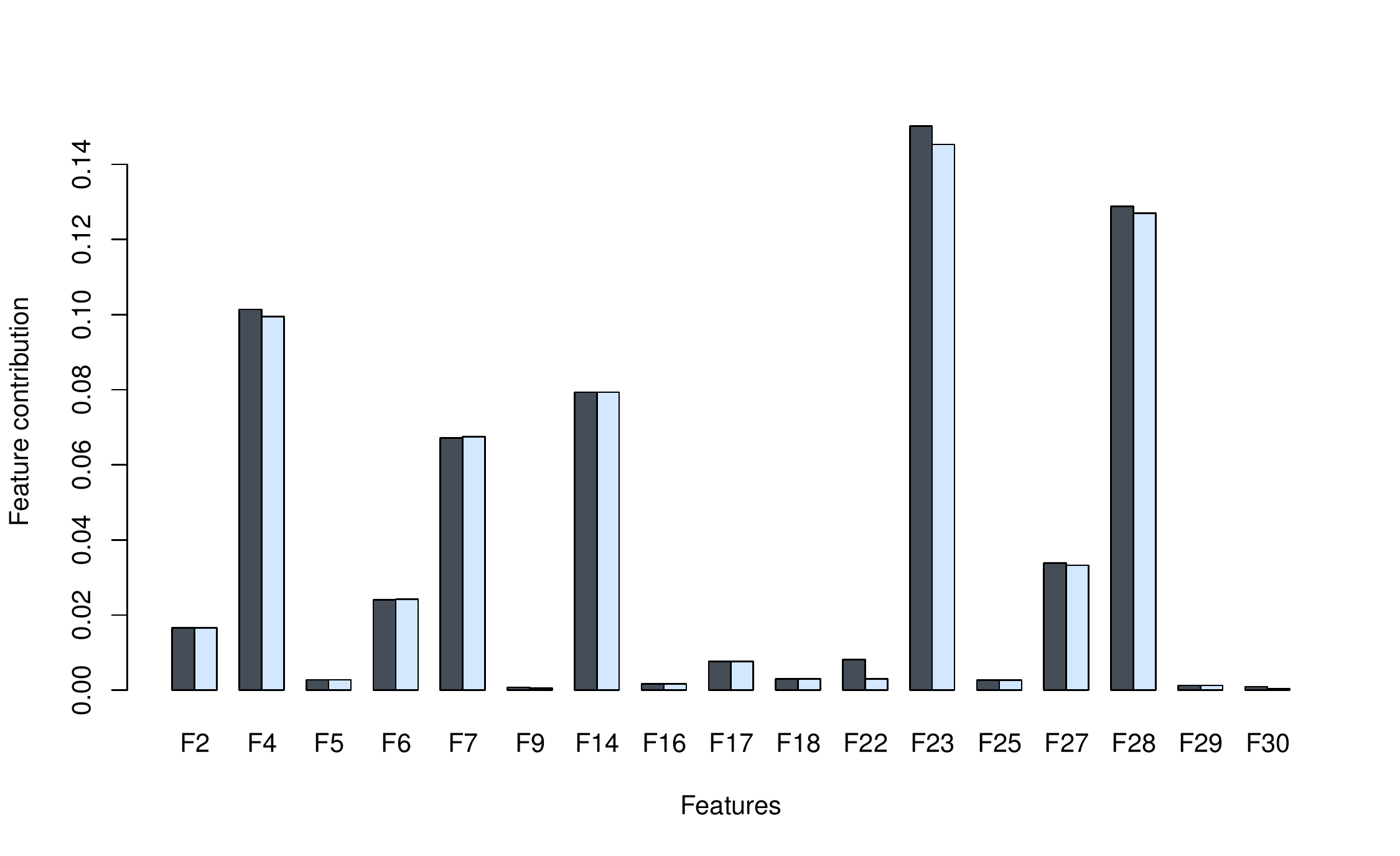}
\caption{}
\label{Fig:full_fc}
\end{subfigure}
\begin{subfigure}[b]{1\textwidth}
\centering
\includegraphics[width=0.6\textwidth, angle=270]{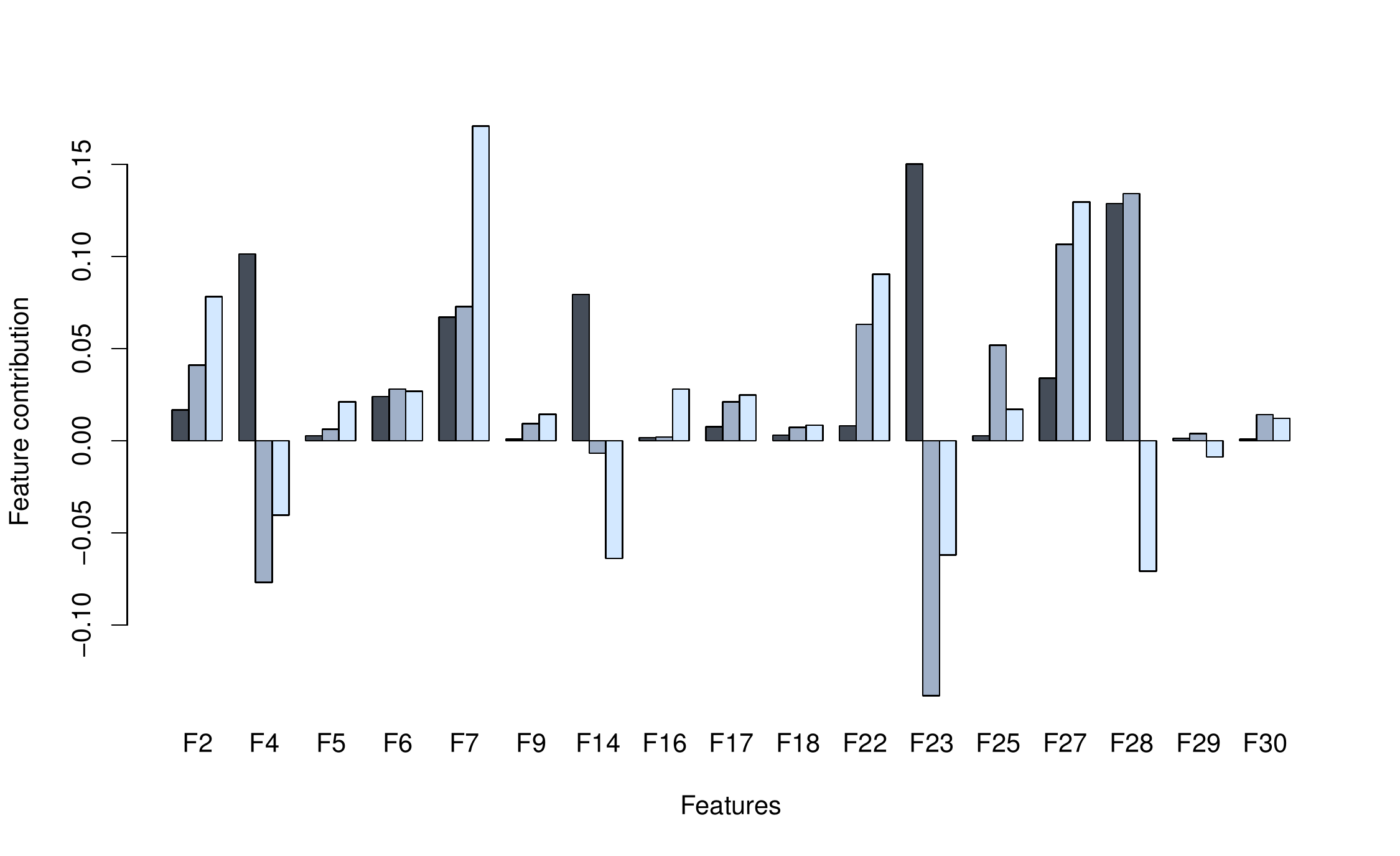}
   \caption{}
\label{Fig:almost_pr}
\end{subfigure}
 \caption{Comparison of the medians of feature contributions (toward class 1) over all instances of class 1 (black bars) with a) feature contributions for instance number 3 (light-grey bars) b) feature contributions for instances number 194 (grey bars) and 537 (light-grey bars) from the BCW Dataset. The fractions of trees voting for class 0 and 1 for these three instances are collected in Table \ref{Tab:full_pr}.}
\end{figure}

Comparison of feature contributions for a particular instance with medians of feature contributions for all instances of one class provides valuable information about the forest's prediction. Take an instance predicted to be in class 1. In a typical case when the large majority of trees votes for class 1 the feature contributions for that instance are very close to the median values (see Figure~\ref{Fig:full_fc}). This happens for around $80\%$ of all instances from the testing dataset predicted to be in class 1. However, when the decision is less unanimous, the analysis of feature contributions may reveal interesting information. As an example, we have chosen instances 194 and 537 (see Table \ref{Tab:full_pr}) which were classified correctly as malignant (class 1) by a majority of trees but with a significant number of trees expressing an opposite view. Figure~\ref{Fig:almost_pr} presents feature contributions for these two instances (grey and light grey bars) against the median values for class 1 (black bars). The largest difference can be seen for the contributions of very significant features F23, F4 and F14: it is highly negative for the two instances under consideration compared to a large positive value commonly found in instances of class 1. Recall that a negative value contributes towards the classification in class 0. There are also three new significant attributes (F2, F22 and F27) that contribute towards the correct classification as well as unusual contributions for features F7 and F28. These newly significant features are judged as only moderately important by both of the variable importance methods in Figure  \ref{Fig:varImp}. It is, therefore, surprising to note that the contribution of these three new features was instrumental in correctly classifying instances 194 and 537 as malignant. This highlights the fact that features which may not generally be important for the model may, nonetheless, be important for classifying specific instances. The approach presented in this paper is able to identify such features, whilst the standard variable importance measures for random forest cannot.
\begin{table}[tb]\centering
\caption{Percentage of trees that vote for each class in RF model for a selection of instances from the BCW Dataset.}
\label{Tab:full_pr}
\begin{tabular}{c|cc}
\hline
Instance Id& benign (class 0) & malignant (class 1)\\
\hline
3 & 0& 1\\
194& 0.298 &0.702\\
537&0.234 &0.766\\
\hline
\end{tabular}
\end{table}

\subsection{Cluster Analysis and Log-likelihood}

The training dataset previously derived for the BCW Dataset was partitioned according to the true class labels. A clustering algorithm implemented in the R package kmeans was run separately for each class. This resulted in the determination of three clusters for class 0 and three clusters for class 1. The structure and size of clusters is presented in Table \ref{tbl:clusters_stats}. Each class has one large cluster: cluster 3 for class 0 and cluster 2 for class 1. Both have a bigger concentration of points around the cluster centre (small average distance) than the remaining clusters. This suggests that there is exactly one core cluster corresponding to a class. This explains the success of the analysis based on the median as the vectors of medians are close to the centres of unique core clusters.

\begin{table}[tbh]
\caption{The structure of clusters for BCW Dataset. For each cluster, the size (the number of training instances) is reported in the left column and the average Euclidean distance from the cluster centre among the training dataset instances belonging to this cluster is displayed in the right column.}
\label{tbl:clusters_stats}\centering
\begin{tabular}{c|cc|cc|cc}
\hline
& \multicolumn{2}{c}{Cluster 1} & \multicolumn{2}{c}{Cluster 2} & \multicolumn{2}{c}{Cluster 3}\\
& size & avg. distance & size & avg. distance & size & avg. distance \\
\hline
class 0 & 12 & 0.220  & 16 & 0.262 & 213 & 0.068\\
class 1 &  20 & 0.241 & 109 & 0.111 & 10 & 0.336\\
\hline
\end{tabular}
\end{table}

\begin{figure}[p]\centering
\begin{subfigure}[b]{0.65\textwidth}
\centering
\includegraphics[width=\textwidth, angle=270]{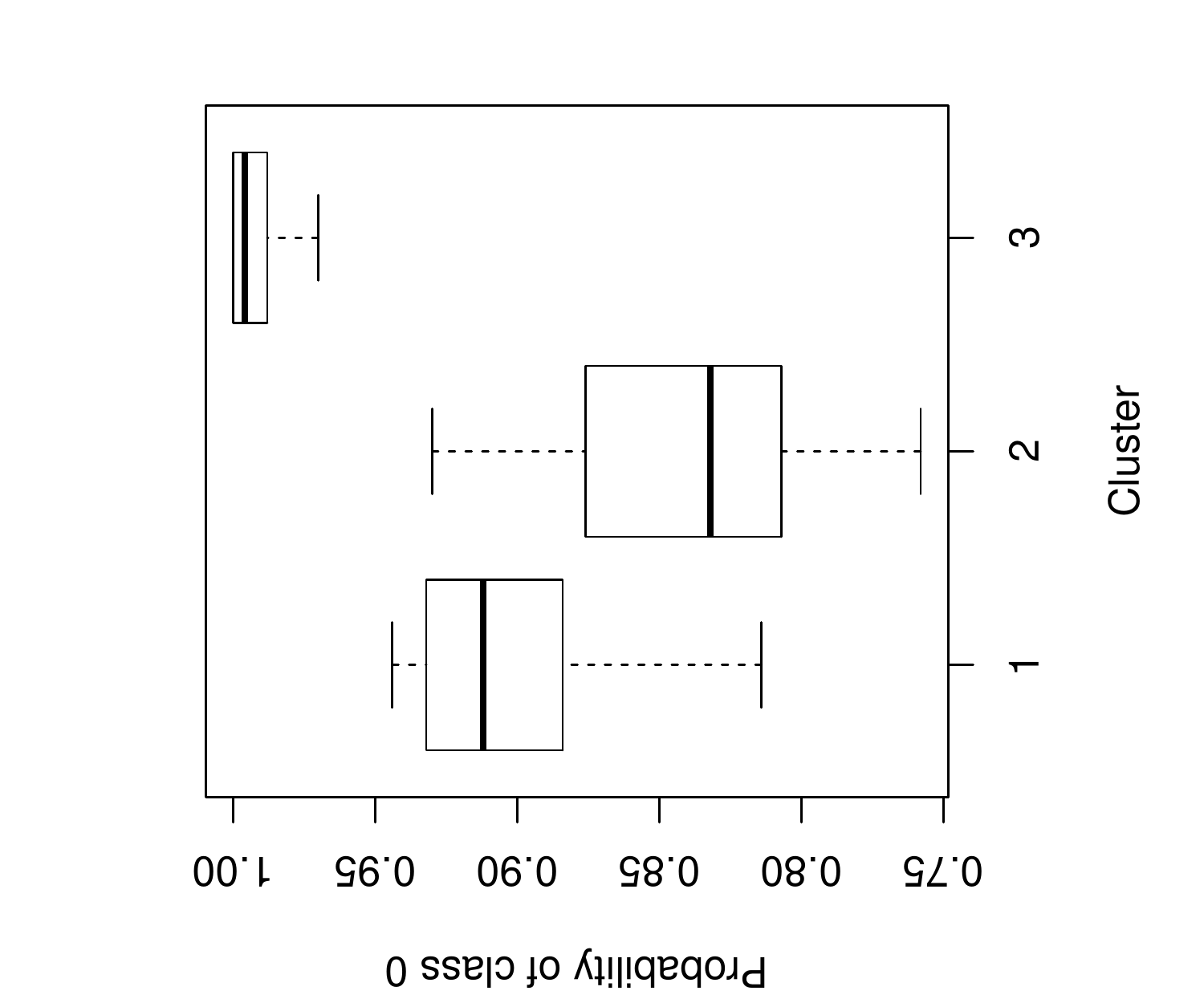}
\caption{Class 0}
\label{Fig:fp0}
\end{subfigure}

 \begin{subfigure}[b]{0.65\textwidth}
\centering
 \includegraphics[width=\textwidth, angle=270]{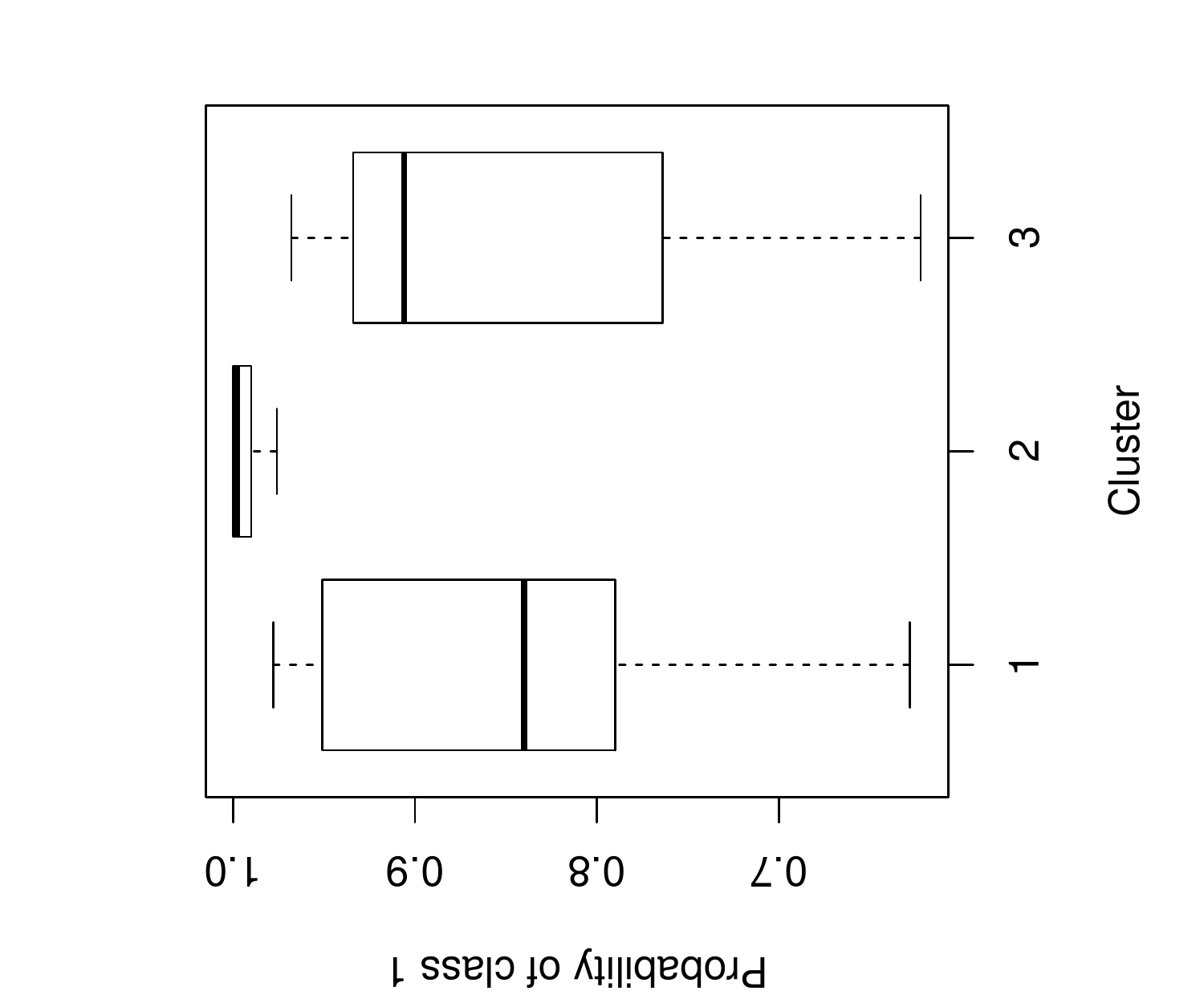}
   \caption{Class 1}
 \label{Fig:fp1}
 \end{subfigure}
    \caption{Fraction of forest trees voting for the correct class in each cluster for training part of the BCW Dataset.}
 \label{Fig:forestprobs}
 \end{figure}

\begin{figure}[p]\centering
\begin{subfigure}[b]{0.6\textwidth}
\includegraphics[width=\textwidth]{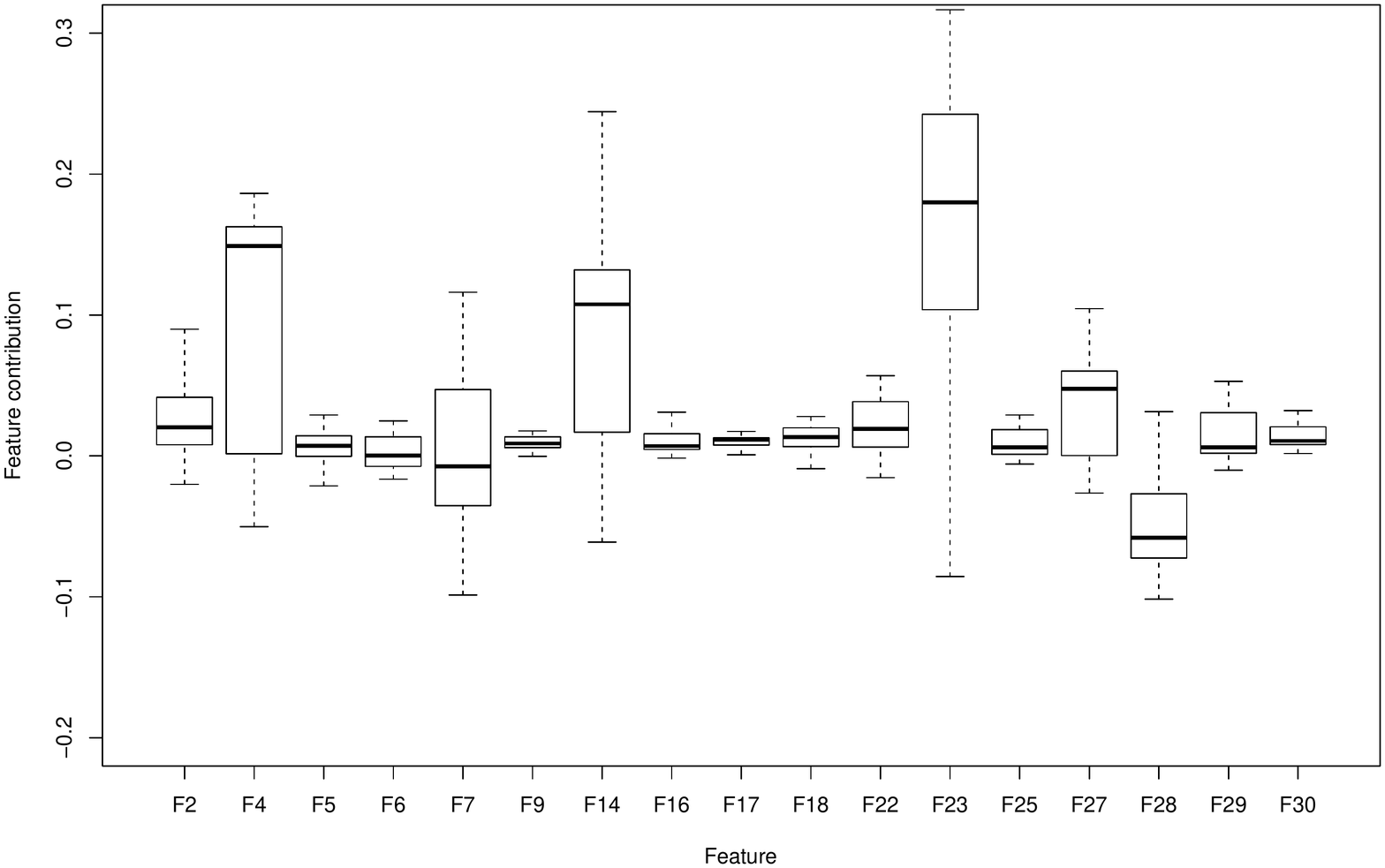}
\caption{Cluster 1 }
\label{Fig:cluster11}
\end{subfigure}
\begin{subfigure}[b]{0.6\textwidth}
\includegraphics[width=\textwidth]{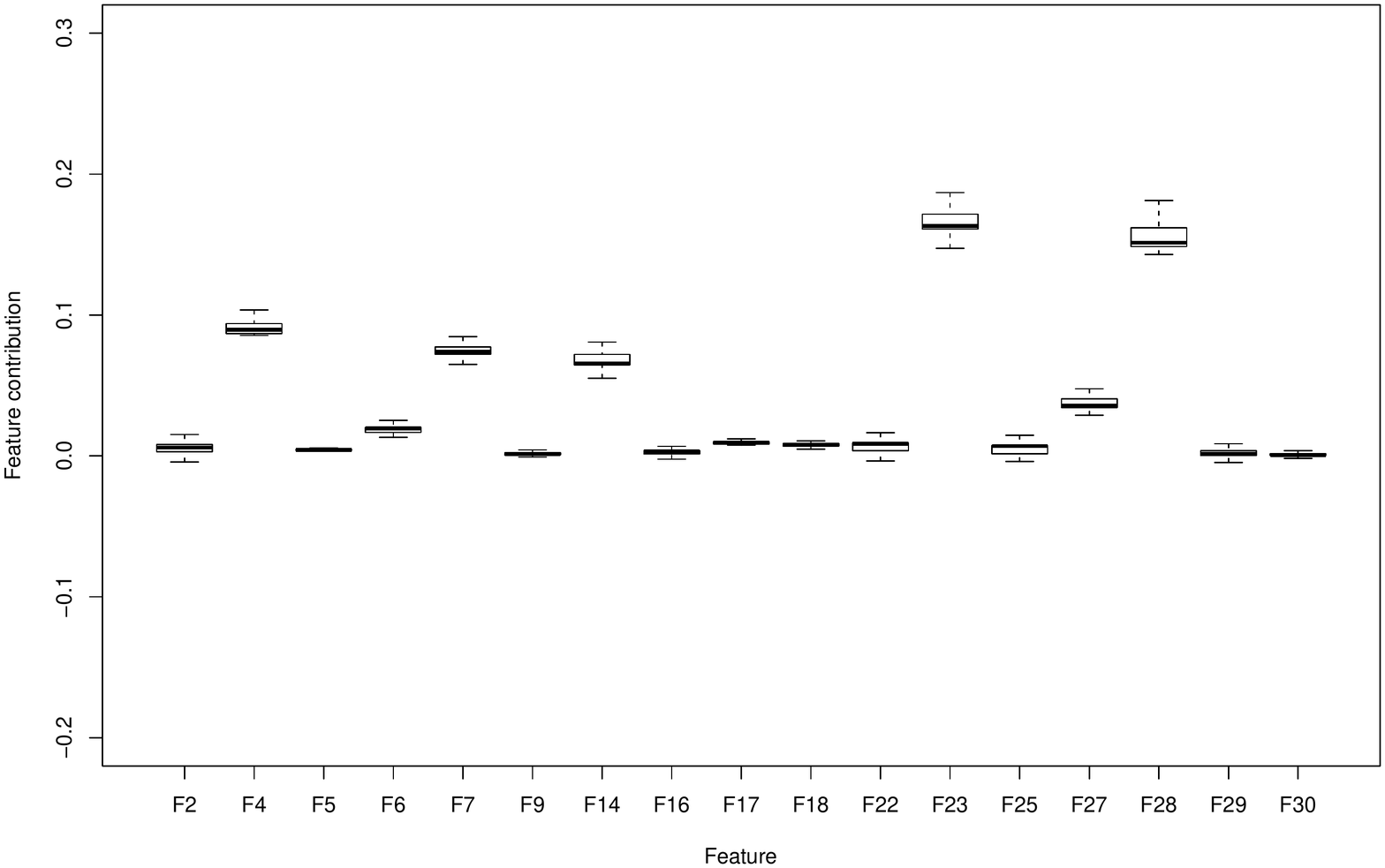}
   \caption{Cluster 2}
\label{Fig:cluster12}
\end{subfigure}
\begin{subfigure}[b]{0.6\textwidth}
\includegraphics[width=\textwidth]{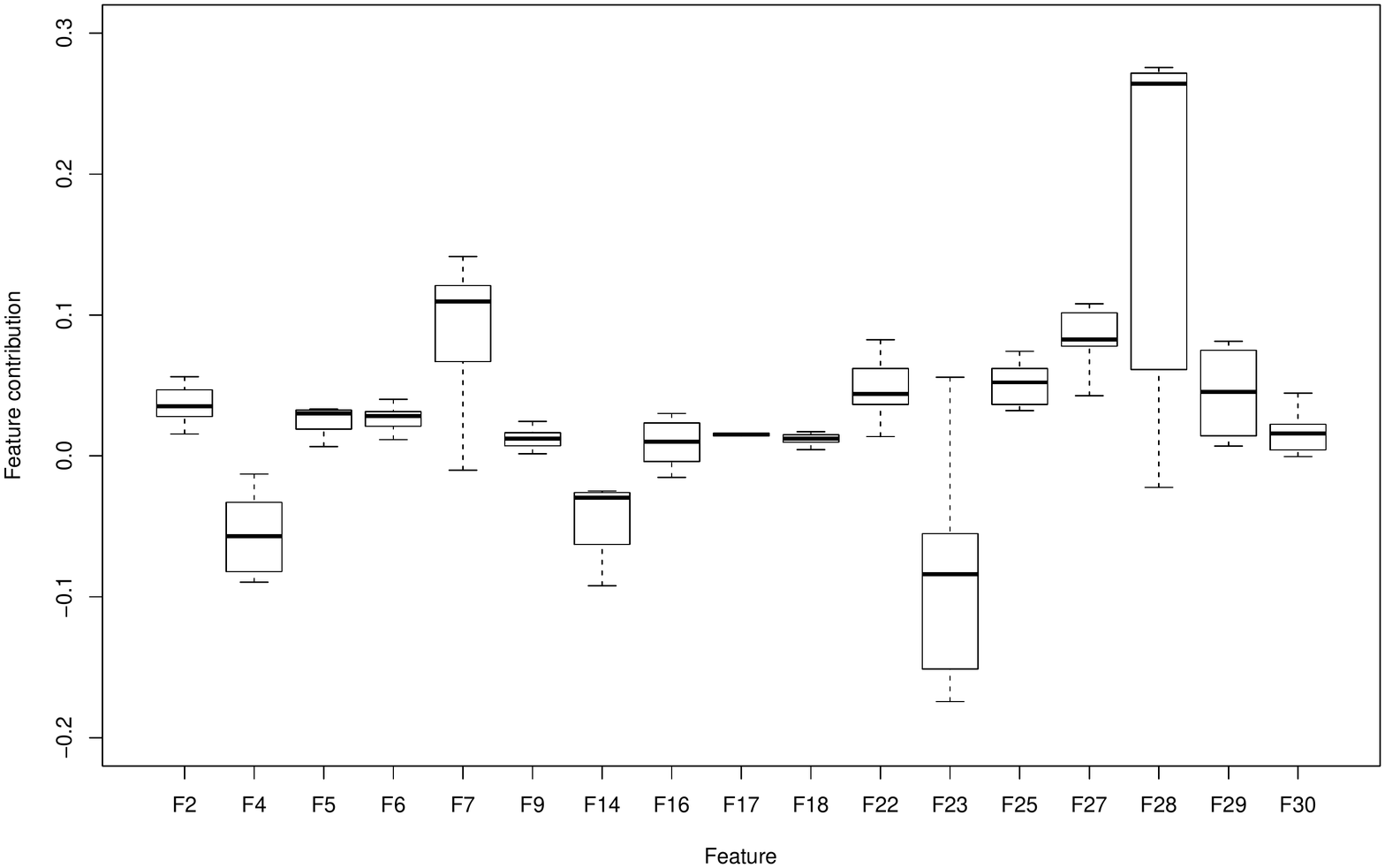}
   \caption{Cluster 3}
\label{Fig:cluster13}
\end{subfigure}
   \caption{Boxplot of feature contributions (towards class 1) for training instances in each of three clusters obtained for class 1.}
\label{Fig:boxplot_cluster1}
\end{figure}

Figure \ref{Fig:forestprobs} lends support to our interpretation of core clusters. The left panel shows the box-plot of the fraction of trees voting for class 0 among training instances belonging to each of the three clusters. A value close to one represents predictions for which the forest is nearly unanimous. This is the case for cluster 3. Two other clusters comprise around $10\%$ of the training instances for which the random forest model happened to be less decisive. A similar pattern can be observed in the case of class 1, see the right panel of the same figure. The unanimity of the forest is observed for the most numerous cluster 2 with other clusters showing lower decisiveness. The reason for this becomes clear once one looks at the variability of feature contributions within each cluster, see Figure \ref{Fig:boxplot_cluster1}. The upper and lower ends of the box corresponds to $25\%$ and $75\%$ quantiles, whereas the whiskers show the full range of the data. Cluster 2 enjoys a minor variability of all the contributions which supports our earlier claims of the similarity of instances (in terms of their feature contributions) in the core class. One can see much higher variability in two remaining clusters showing that the forest used different features as evidence to classify instances in each of these clusters. Although in cluster 2 all contributions were positive, in clusters 1 and 3 there are features with negative contributions. Recall that a negative value of a feature contribution provides evidence against being in the corresponding class, here class 1.

\begin{figure}[tbh]
\centering
\includegraphics[width=0.8\textwidth, angle=0]{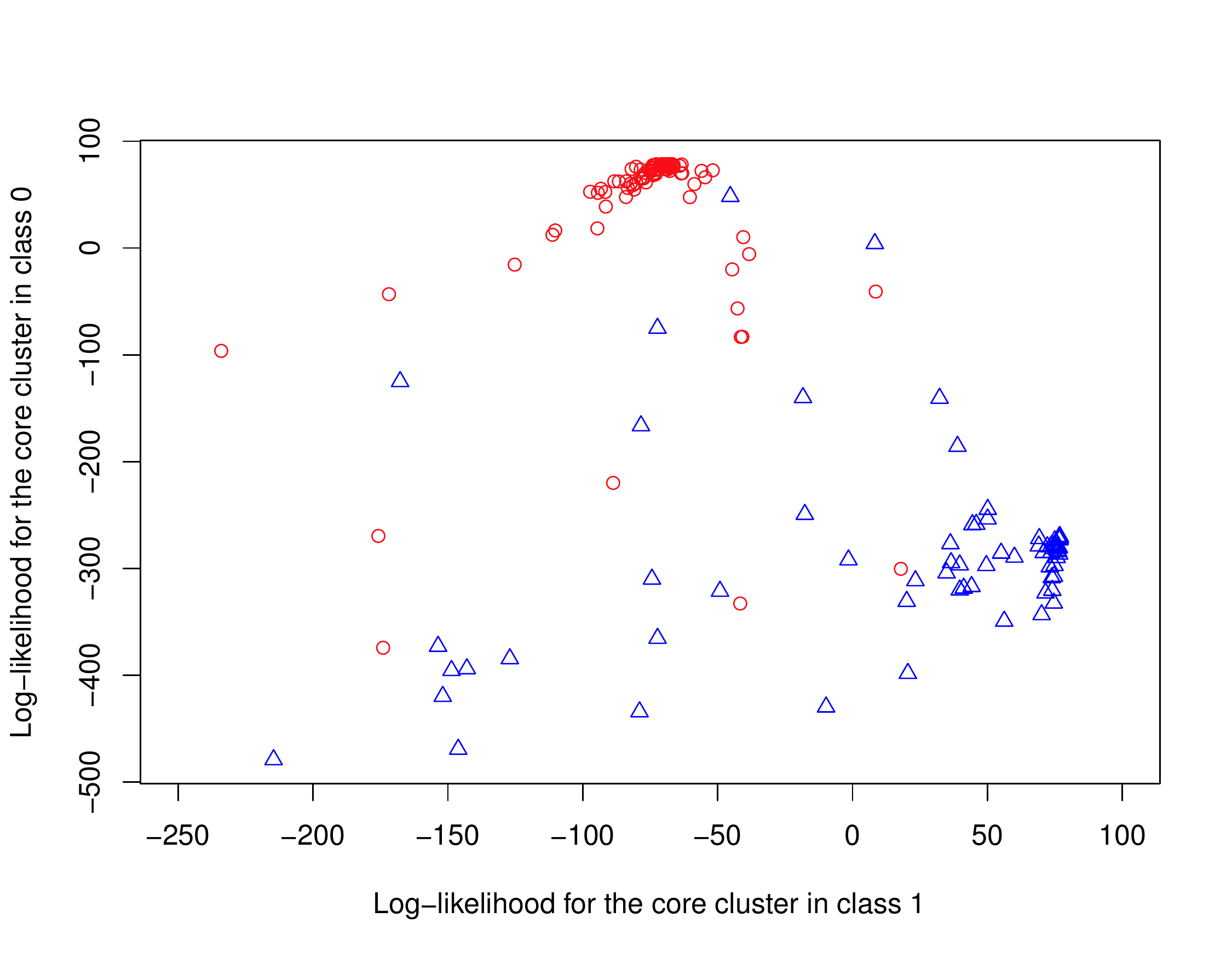}
    \caption{Log-likelihoods for belonging to the core cluster in class 0 (vertical axis) and class 1 (horizontal axis) for the testing dataset in BCW. Circles correspond to instances of class 0 while triangles denote instances of class 1.}
 \label{Fig:loglikelihood}
 \end{figure}

Based on the observation that clusters correspond to a particular decision-making route for the random forest model, we introduced the log-likelihood as a way to assess the distance of a given instance from the cluster centre, or, in a probabilistic interpretation, to compute the likelihood\footnote{The likelihood is obtained by applying the exponential function to the log-likelihood.} that the instance belongs to the given cluster.  It should however be clarified that one cannot compare the likelihood for the core cluster in class 0 with the likelihood for the core cluster in class 1. The likelihood can only be used for comparisons within one cluster: having two instances we can say which one is more likely to belong to a given cluster. By comparing it to a typical likelihood for training instances in a given cluster we can further draw conclusions about how well an instance fits that cluster. Figure \ref{Fig:loglikelihood} presents the log-likelihoods for the two core clusters (one for each class) for instances from the testing dataset. Shapes are used to mark instances belonging to each class: circles for class 0 and triangles for class 1. Notice that likelihoods provide a very good split between classes: instances belonging to class 0 have a high log-likelihood for the core cluster of class 0 and rather low log-likelihood for the core cluster of class 1. And vice-versa for instances of class 1.

\subsection{Iris Dataset}

\begin{figure}[thb]\centering
\includegraphics[width=0.7\textwidth]{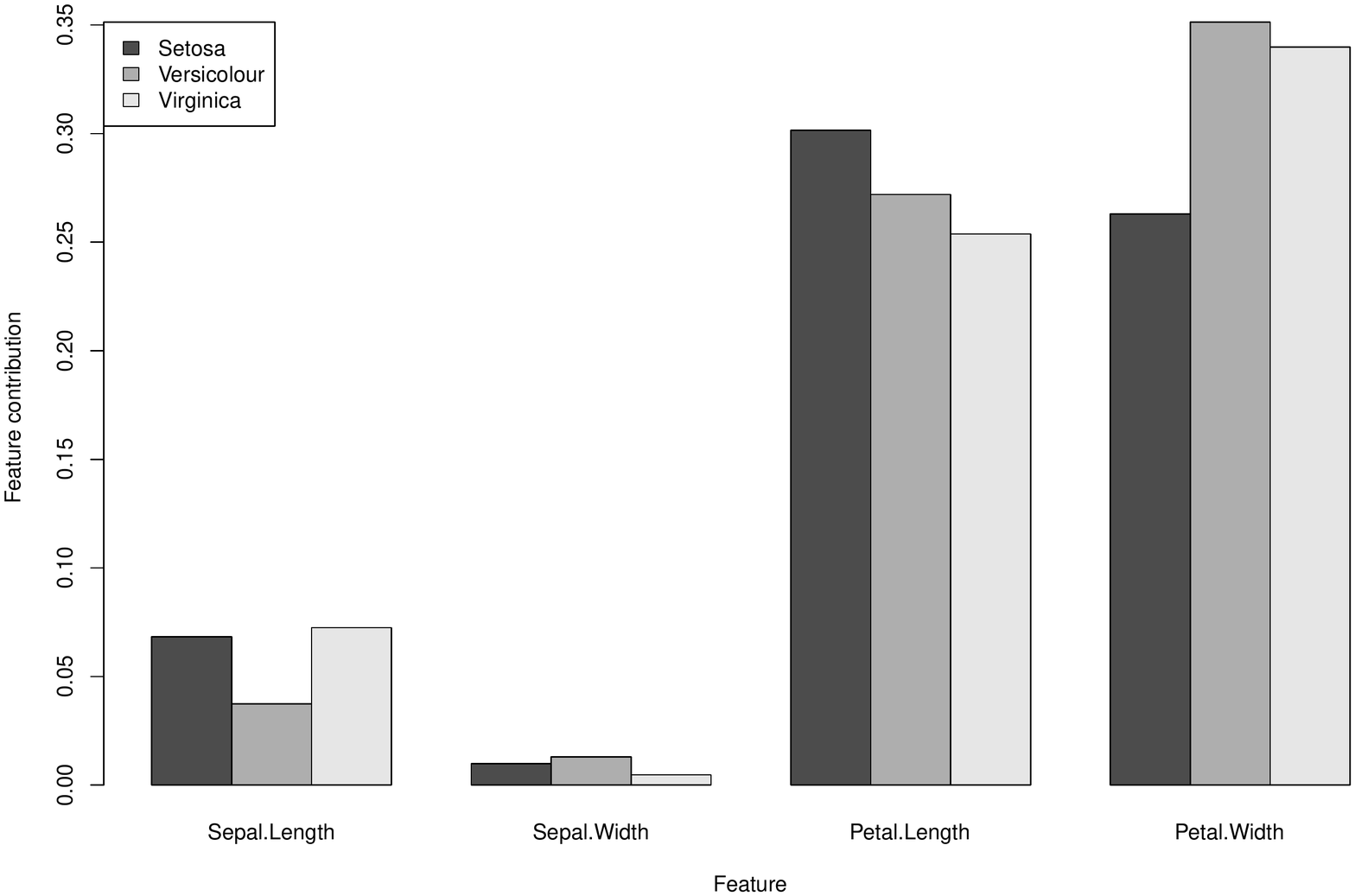}
\caption{Medians of feature contributions for each class for the UCI Iris Dataset.}
\label{Fig:median_iris}
\end{figure}

In this section we use the UCI Iris Dataset~\cite{Iris} to demonstrate interpretability of  feature contributions for multi-classification models. We generated a random forest model on 100 randomly selected instances. The remaining 50 instances were used to assess the accuracy of the model: 47 out of 50 instances were correctly classified. Then we applied our approach for determining the feature contributions for the generated model. Figure~\ref{Fig:median_iris} presents medians of feature contributions for each of the three classes. In contrast to the binary classification case, the medians are positive for all classes. A positive feature contribution for a given class means that the value of this feature directs the forest towards assigning this class. A negative value points towards the other classes.

\begin{table}[b]\centering
\caption{Feature contributions towards predicted classes for selected instances from the UCI Iris Dataset.}
\label{table:iris}
 \begin{tabular}{c|cc|cc}
\multirow{2}{*}{Instance} & \multicolumn{2}{c|}{Sepal} & \multicolumn{2}{c}{Petal}\\
& Length & Width & Length & Width \\
\hline
120  & 0.059   & 0.014  & 0.053  &  0.448\\
150  & -0.097 & 0.035 & 0.259  &  0.339\\
\hline
 \end{tabular}
\end{table}

Feature contributions provide valuable information about the reliability of random forest predictions for a particular instance. It is commonly assumed that the more trees voting for a particular class, the higher the chance that the forest decision is correct. We argue that the analysis of feature contributions offers a more refined picture. As an example, take two instances: 120 and 150. The first one was classified in class Versicolour (88\% of trees voted for this class). The second one was assigned class Virginica with 86\% of trees voting for this class. We are, therefore, tempted to trust both of these predictions to the same extent. Table \ref{table:iris} collects feature contributions for these instances towards their predicted classes. Recall that the highest contribution to the decision is commonly attributed to features 3 (Petal.Length) and 4 (Petal Width), see Figure \ref{Fig:median_iris}.
These features also make the highest contributions to the predicted class for instance 150. The indecisiveness of the forest may stem from an unusual value for the feature 1 (Sepal.Length) which points towards a different class. In contrast, the instance 120 shows standard (low) contributions of the first two features and unusual contributions of the last two features: very low for feature 3 and high for feature 4. Recall that features 3 and 4 tend to contribute most to the forest's decision (see Figure~\ref{Fig:median_iris}) with values between 0.25 and 0.35. The low value for feature 3 is non-standard for its predicted class, which increases the chance of it being wrongly classified. Indeed, both instances belong to class Virginica while the forest classified the instance 120 wrongly as class Versicolour and the instance 150 correctly as class Virginica.

\begin{figure}[tb!]\centering
\includegraphics[width=0.9\textwidth]{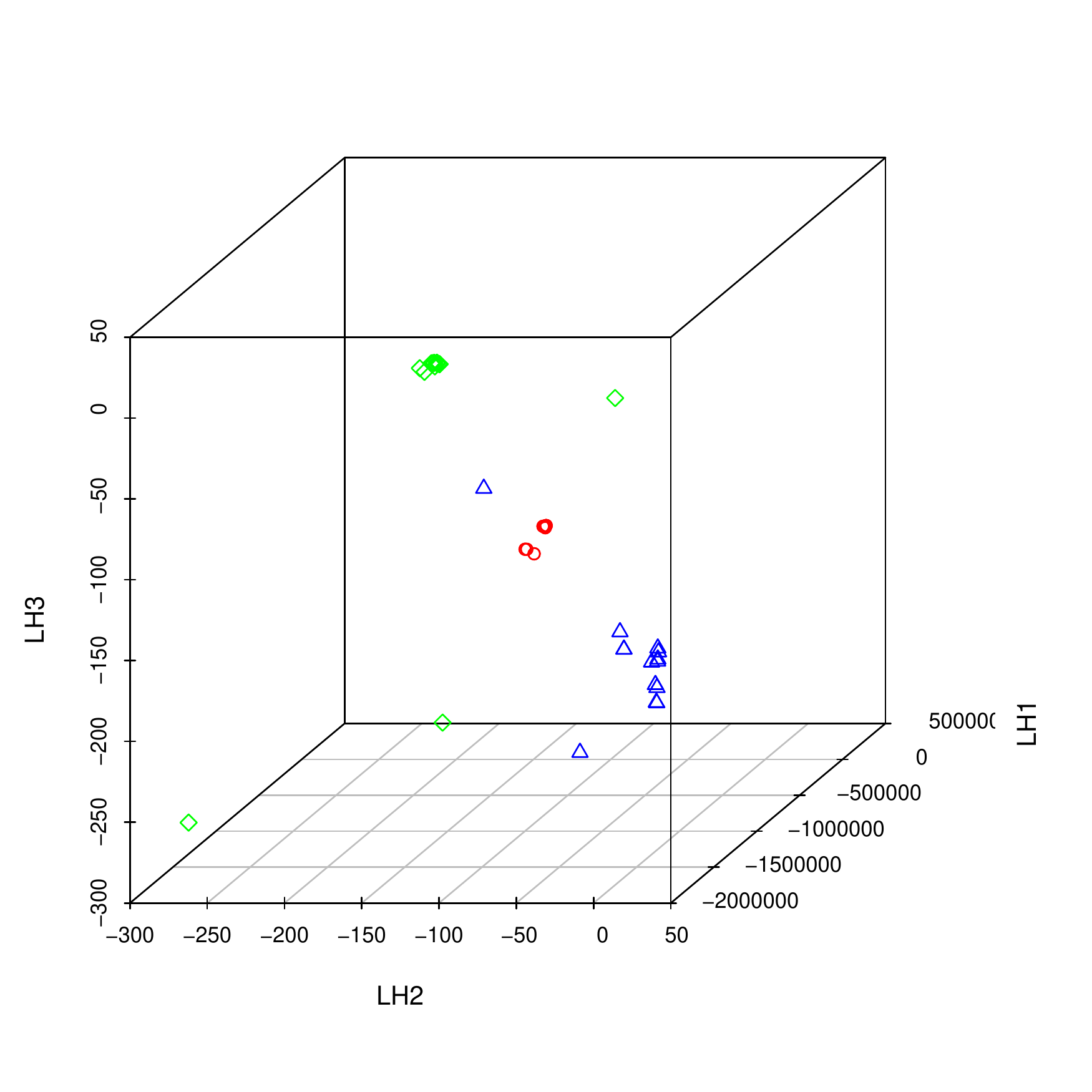}
\caption{Log-likelihoods for all instances in UCI Iris Dataset towards core clusters for each class. {Circles represent the Setosa class, triangles represent Versicolour  and diamonds represent the Virginica class. Points corresponding to the same class tend to group together and there are only a few instances that are far from their cores.} }
\label{Fig:3d_iris}
\end{figure}

The cluster analysis of feature contributions for the UCI Iris Dataset revealed that it is sufficient to consider only two clusters for each class. Cluster sizes are $5$ and $45$ for class Setosa, $4$ and $46$ for class Versicolour and $5$ and $44$ for class Virginica. Core clusters were straightforward to determine: for each class, the largest of the two clusters was selected as the core cluster. Figure \ref{Fig:3d_iris} displays an analysis of log-likelihoods for all instances in the dataset. For every instance, we computed feature contributions towards each class and calculated log-likelihoods of being in the core clusters of the respective classes. On the graph, each point represents one instance. The coordinate LH1 is the log-likelihood for the core cluster of class Setosa, the coordinate LH2 is the log-likelihood for the core cluster of class Versicolour and the coordinate LH3 is the log-likelihood for the core cluster of class Virginica. Shapes of points show the true classification: class Setosa is represented by circles, Versicolour by triangles and Virginica by diamonds. Notice that points corresponding to instances of the same class tend to group together. This can be interpreted as the existence of coherent patterns in the reasoning of the random forest model.

\subsection{Robustness Analysis}

For the validity of the study of feature contributions, it is crucial that the results are not artefacts of one particular realization of a random forest model but that they convey actual information held by the data. We therefore propose a method for robustness analysis of feature contributions. We will use the UCI Breast Cancer Wisconsin Dataset studied in Subsection \ref{subsec:empirics_breast} as an example.

\begin{figure}[p]\centering
\begin{subfigure}[b]{\textwidth}\centering
\includegraphics[width=0.6\textwidth]{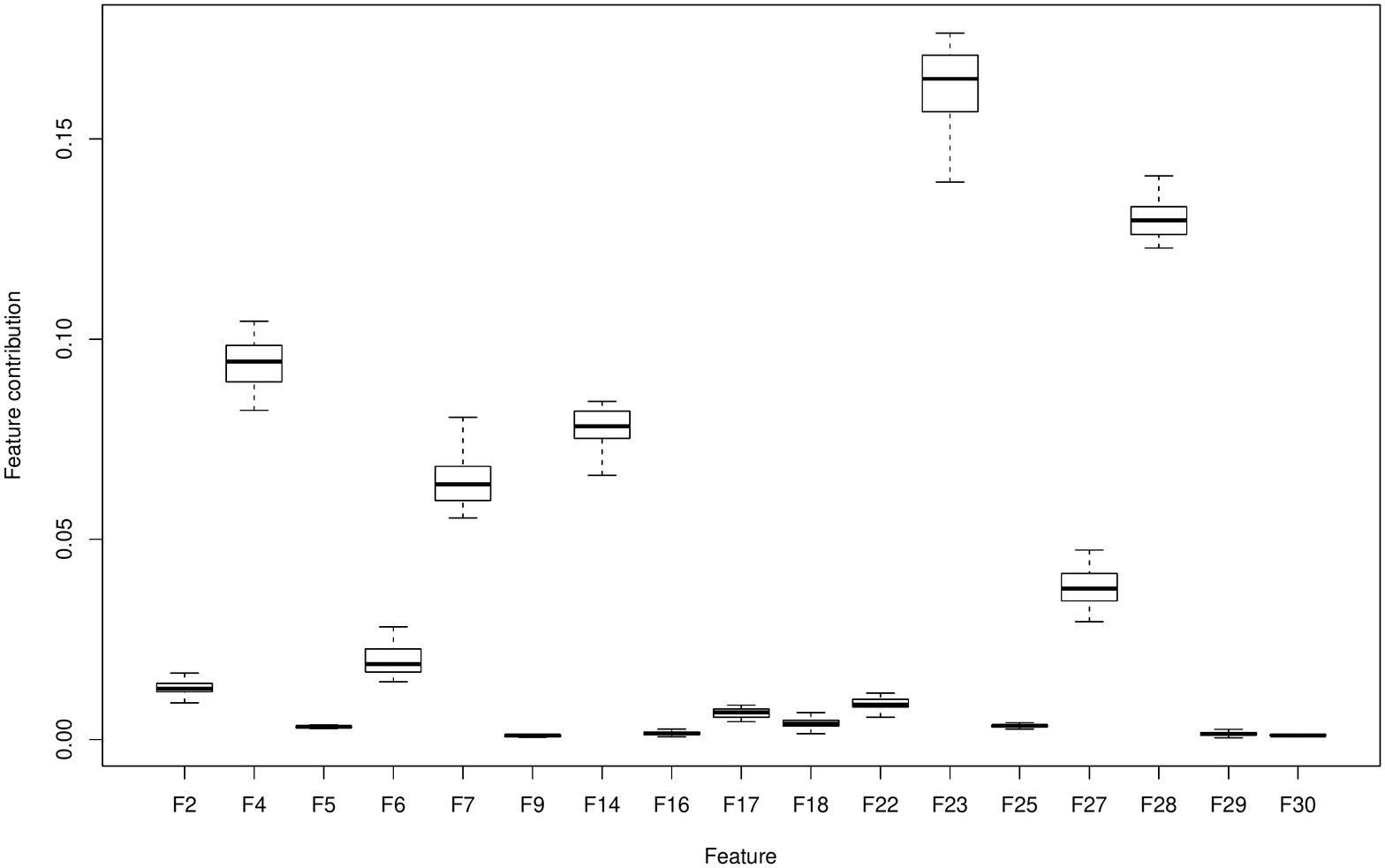}
   \caption{(a) Medians of feature contributions for training datasets}
\label{Fig:robustness_train}
\end{subfigure}
\begin{subfigure}[b]{\textwidth}\centering
\includegraphics[width=0.6\textwidth]{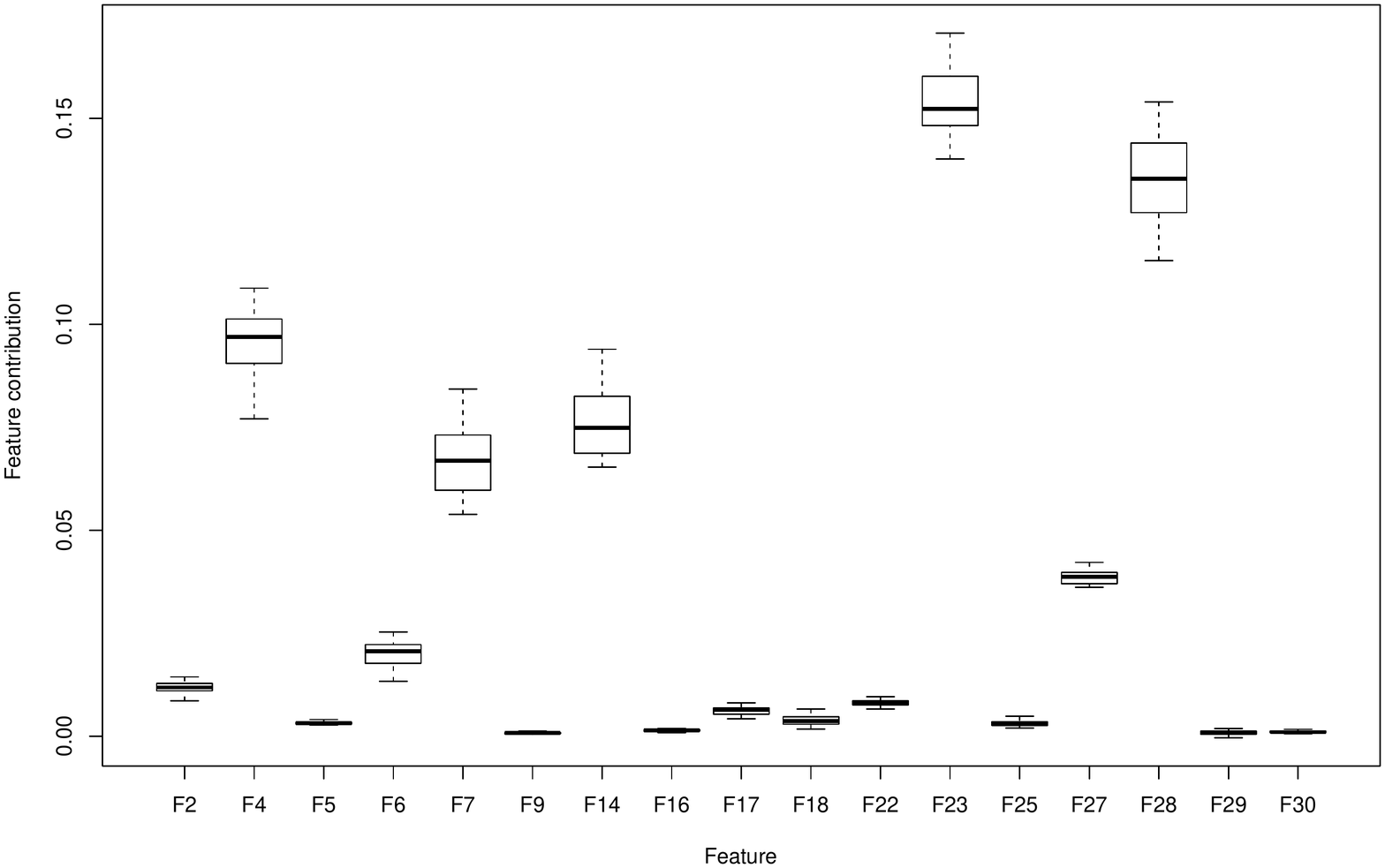}
   \caption{(b) Medians of feature contributions for testing datasets}
\label{Fig:robustness_test}
\end{subfigure}
\begin{subfigure}[b]{\textwidth}\centering
\includegraphics[width=0.6\textwidth]{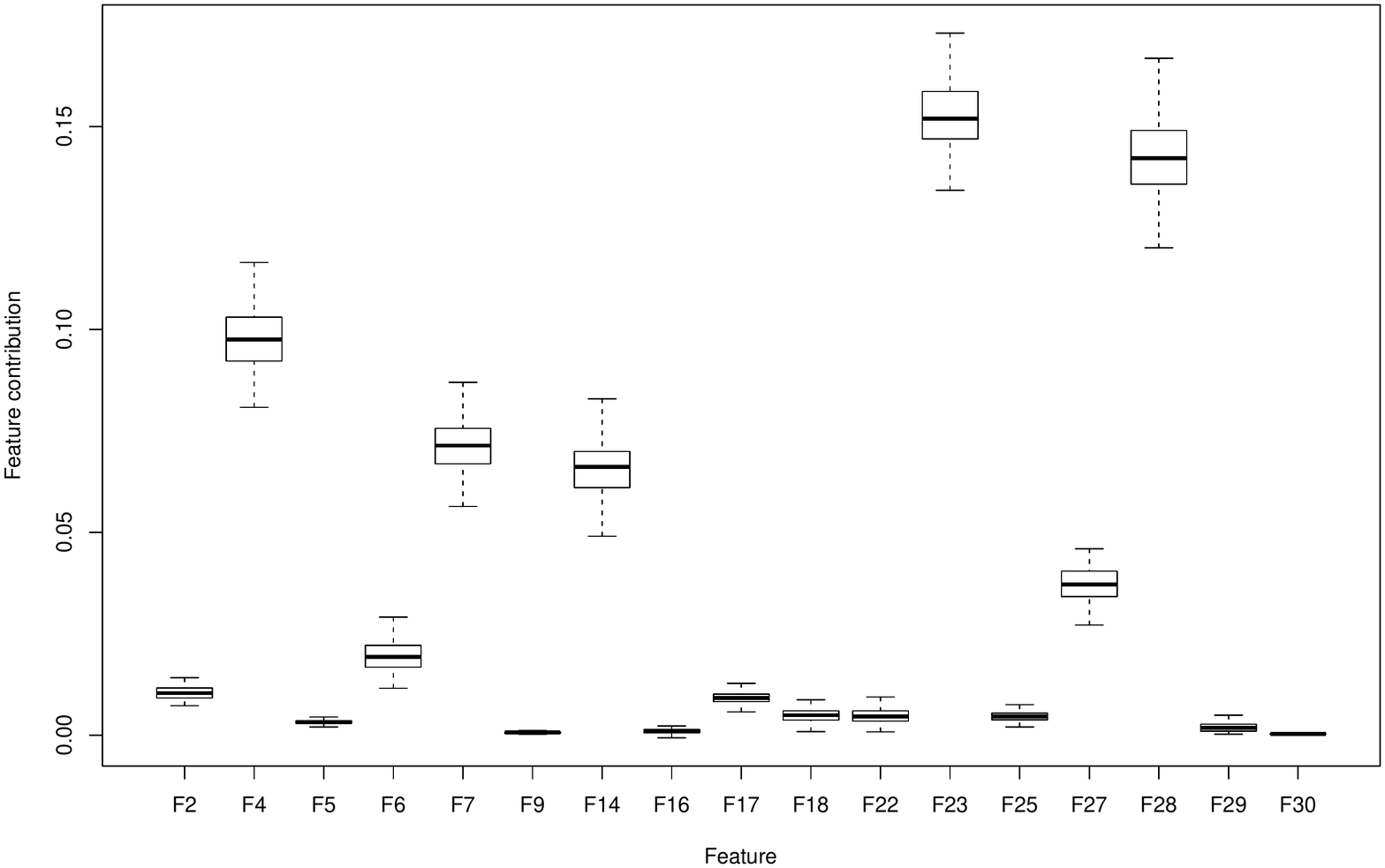}
   \caption{(c) Feature contributions for an unseen instance}
\label{Fig:robustness_sample3}
\end{subfigure}
   \caption{Feature contributions towards class 1 for 100 random forest models for the BCW dataset.}
\label{Fig:robustness}
\end{figure}

We removed instance number 3 from the original dataset to allow us to perform tests with an unseen instance. We generated 100 random forest models with 500 trees with each model built using an independent randomly generated training set with $379 \approx 2/3 \cdot 568$ instances. The rest of the dataset for each model was used for its validation. The average model accuracy was 0.963. For each generated model, we collected medians of feature contributions separately for training and testing datasets and each class. The variation of these quantities over models for class 1 and the training dataset are presented using a box plot in Figure~\ref{Fig:robustness_train}. The top of the box is the $75\%$ quantile, the bottom is the $25\%$ quantile, while the bold line in the middle is the median (recalling that this is the median of the median feature contributions across multiple models). Whiskers show the extent of minimal and maximal values for each feature contribution. Notice that the variation between simulations is moderate and conclusions drawn for one realization of the random forest model in Subsection \ref{subsec:empirics_breast} would hold for each of the generated 100 random forest models.

A testing dataset contains those instances that do not take part in the model generation. One can, therefore, expect more errors in the classification of the forest, which, in effect, should imply lower stability of the calculated feature contributions. Indeed, the box plot presented in Figure~\ref{Fig:robustness_test} shows a slight tendency towards increased variability of the feature contributions when compared to Figure \ref{Fig:robustness_train}. However, these results are qualitatively on a par with those obtained on the training datasets. We can, therefore, conclude that feature contributions computed for a new (unseen) instance provide reliable information. We further tested this hypothesis by computing feature contributions for instance number 3 that did not take part in the generation of models. The statistics for feature contributions for this instance over 100 random forest models are shown in Figure~\ref{Fig:robustness_sample3}. Similar results were obtained for other instances.

\section{Conclusions}\label{sec:conclusions}

Feature contributions provide a novel approach towards black-box model interpretation. They measure the influence of variables/features on the
prediction outcome and provide explanations as to why a model makes a particular decision. In this work, we extended the feature contribution method of \cite{Kuzmin:2011} to random forest classification models and we proposed three techniques (median, cluster analysis and log-likelihood) for finding patterns in the random forest's use of available features. Using UCI benchmark datasets we showed the robustness of the proposed methodology. We also demonstrated how feature contributions can be applied to understand the dependence between instance characteristics and their predicted classification and to assess the reliability of the prediction. The relation between feature contributions and standard variable importance measures was also investigated. The software used in the empirical analysis was implemented in R as an add-on for the \texttt{randomForest} package and is currently being prepared for submission to CRAN~\cite{CRAN} under the name \texttt{rfFC}.

\medskip

\textbf{Acknowledgements.} This work is partially supported by BBSRC and Syngenta Ltd through the Industrial CASE Studentship Grant (No. BB/H530854/1).


\bibliographystyle{plain}
\bibliography{feature_contrib}

\begin{thebibliography}{10}

\bibitem{BreastCancer}
{Breast Cancer Wisconsin Diagnostic dataset}.
\newblock
  \url{http://archive.ics.uci.edu/ml/datasets/Breast+Cancer+Wisconsin+\%28Diag%
nostic\%29}.

\bibitem{CRAN}
{CRAN - The Comprehensive R Archive Network}.
\newblock \url{http://cran.r-project.org/}.

\bibitem{Iris}
Iris dataset.
\newblock \url{http://archive.ics.uci.edu/ml/datasets/Iris}.

\bibitem{Baehrens2010}
David Baehrens, Timon Schroeter, Stefan Harmeling, Motoaki Kawanabe, Katja
  Hansen, and Klaus-Robert Muller.
\newblock How to explain individual classification decisions.
\newblock {\em Journal of Machine Learning Research}, 11:1803--1831, 2010.

\bibitem{Breiman:2001}
Leo Breiman.
\newblock Random forests.
\newblock {\em Machine Learning}, 45(1):5--32, 2001.

\bibitem{Breiman:2008}
Leo Breiman and Adele Cutler.
\newblock Random forests.
\newblock \url{http://www.stat.berkeley.edu/~breiman/RandomForests/}, 2008.

\bibitem{Breiman:1984}
Leo Breiman, J.~H. Friedman, R.~A. Olshen, and C.~J. Stone.
\newblock {\em Classification and regression trees}.
\newblock Monterey, CA: Wadsworth \& Brooks/Cole Advanced Books \& Software,
  1984.

\bibitem{Carlsson2009}
Lars Carlsson, Ernst~Ahlberg Helgee, and Scott Boyer.
\newblock Interpretation of nonlinear qsar models applied to ames mutagenicity
  data.
\newblock {\em Journal of Chemical Information and Modeling},
  49(11):2551--2558, 2009.

\bibitem{Cormen:2001}
Thomas~H. Cormen, Clifford Stein, Ronald~L. Rivest, and Charles~E. Leiserson.
\newblock {\em Introduction to Algorithms}.
\newblock McGraw-Hill Higher Education, 2nd edition, 2001.

\bibitem{Hand:2001}
David~J. Hand, Padhraic Smyth, and Heikki Mannila.
\newblock {\em Principles of data mining}.
\newblock MIT Press, Cambridge, MA, USA, 2001.

\bibitem{Hansen2011}
Katja Hansen, David Baehrens, Timon Schroeter, Matthias Rupp, and Klaus-Robert
  Muller.
\newblock Visual interpretation of kernel-based prediction models.
\newblock {\em Molecular Informatics}, 30(9):817--826, 2011.

\bibitem{Kuzmin:2011}
Victor~E. Kuz'min, Pavel~G. Polishchuk, Anatoly~G. Artemenko, and Sergey~A.
  Andronati.
\newblock Interpretation of qsar models based on random forest methods.
\newblock {\em Molecular Informatics}, 30(6-7):593--603, 2011.

\bibitem{Liaw:2002}
Andy Liaw and Matthew Wiener.
\newblock Classification and regression by randomforest.
\newblock {\em R News}, 2(3):18--22, 2002.

\bibitem{Rajaraman:2012}
Anand Rajaraman and Jeffrey~D. Ullman.
\newblock {\em Mining of Massive Datasets}.
\newblock Cambridge University Press, 2012.

\bibitem{Rosenbaum2011}
Lars Rosenbaum, Georg Hinselmann, Andreas Jahn, and Andreas Zell.
\newblock Interpreting linear support vector machine models with heat map
  molecule coloring.
\newblock {\em Journal of Cheminformatics}, 3(1):11, 2011.

\bibitem{Strobl2008}
Carolin Strobl, Anne-Laure Boulesteix, Thomas Kneib, Thomas Augustin, and Achim
  Zeileis.
\newblock Conditional variable importance for random forests.
\newblock {\em BMC Bioinformatics}, 9(1):307, 2008.

\bibitem{Tropsha:2010}
Alexander Tropsha.
\newblock Best practices for qsar model development, validation, and
  exploitation.
\newblock {\em Molecular Informatics}, 29(6-7):476--488, 2010.

\end{thebibliography}

\end{document}